\documentclass[prodmode,acmtap]{acmlarge}

\acmVolume{0}
\acmNumber{0}
\acmArticle{0}
\articleSeq{0}
\acmYear{0}
\acmMonth{0}

\usepackage[ruled]{algorithm2e}
\usepackage{natbib}
\usepackage{graphicx,xcolor}
\usepackage{booktabs}
\usepackage{lipsum}
\usepackage{microtype}
\usepackage{url}
\usepackage[final]{pdfpages}
\usepackage{fancyhdr}
\usepackage{listings}
\usepackage{hyperref}
\usepackage{color}
\usepackage{tikz}
\usepackage{caption}
\usepackage{lscape}
\SetAlFnt{\algofont}
\SetAlCapFnt{\algofont}
\SetAlCapNameFnt{\algofont}
\SetAlCapHSkip{0pt}
\IncMargin{-\parindent}
\renewcommand{\algorithmcfname}{ALGORITHM}

\title{Automated text summarisation and evidence-based medicine: A survey of two domains}
\author{ABEED SARKER and DIEGO MOLL\'A \affil{Macquarie University}
C\'ECILE PARIS
\affil{CSIRO Australia}
}

\begin{abstract}

The practice of evidence-based medicine (EBM) urges medical practitioners to utilise the latest research evidence when making clinical decisions. Because of the massive and growing volume of published research on various medical topics, practitioners often find themselves overloaded with information. As such, natural language processing research has recently commenced exploring techniques for performing medical domain-specific automated text summarisation (ATS) techniques--- targeted towards the task of condensing large medical texts. However, the development of effective summarisation techniques for this task requires cross-domain knowledge. We present a survey of EBM, the domain-specific needs for EBM, automated summarisation techniques, and how they have been applied hitherto. We envision that this survey will serve as a first resource for the development of future operational text summarisation techniques for EBM.

\end{abstract}



\begin{document}


\maketitle

\section{Introduction}
Evidence-based medicine (EBM) promotes a form of medical practice which requires the incorporation of the current, best research evidence available on a topic by medical practitioners, combine it with their expertise, and the patients' preferences, when making medical decisions. However, the amount of available medical research literature is massive, presenting practitioners with the problem of information overload. As such, there has been increasing interest in recent times on research targeted towards the development of post-retrieval systems that can summarise information based on the needs of medical practitioners. Text summarisation is a field of research within the domain of natural language processing (NLP), and its aim is to generate summaries from large volumes of text. While research on open-domain text summarisation has seen significant progress, its application to problems involving more complex forms of natural language, such as those in the medical domain, is still very much in its infancy \cite{sarker16summ}. This can be attributed to the fact that systems attempting to summarise information in such complex domains require the incorporation of vast amounts of domain-specific knowledge, which practitioners acquire through years of experience, and integrate that knowledge with NLP techniques targeted towards summarisation of information. Thus, NLP researchers intending to develop end-to-end summarisation systems to support EBM require an in-depth understanding of the practice as well as techniques for automated summarisation. The intent of this survey is to bring together information from the distinct domains of EBM and ATS, and to provide a single resource for researchers attempting to develop automated summarisation systems addressing the needs of this practice. 

In this paper, we review the state-of-the-art for EBM. We first provide a detailed description of EBM, its goals, the obstacles faced by practitioners, and a brief overview of how various NLP techniques can aid the practice. In the second part of this paper, we focus on ATS. We first provide a brief review of generic summarisation techniques, and then present a discussion of the state-of-the-art in summarisation/question-answering technologies in this domain, pointing out the gaps in research that need to be filled to implement end-to-end systems. We then provide a relatively detailed discussion of summarisation approaches that have been applied to the medical domain, and analyse some recent summarisation approaches in detail. In particular, we attempt to answer the following questions:

\begin{itemize}
\item What is EBM and what are the major obstacles hindering EBM practice?
\item What are the characteristics of text in the medical domain?
\item What tools and resources are available for text processing in the medical domain? 
\item What is ATS and what is its relevance to EBM?
\item What is the current state of research in ATS, particularly for the medical domain?
\item How can ATS approaches for EBM be evaluated?
\end{itemize}

The rest of the article is organised as follows. In Section 2 we provide a detailed overview of EBM, the technological needs of the practice, and the challenges associated with it. We also discuss the tools and resources that are currently available to EBM practitioners. We review the literature associated with ATS in Section 3 and discuss some key techniques that have been applied in the past. In Section 4, we focus our review on question answering and summarisation techniques specific to the medical domain, including current state-of-the-art in evidence-based text summarisation.  In Section 5, we briefly review some evaluation techniques for ATS and discuss their applicability to EBM. We conclude the paper in Section 6, and summarise the key discoveries from our survey.

\section{Evidence-based Medicine}
The phrase `evidence-based medicine' was defined initially as \emph{``a process of turning clinical problems into questions and then systematically locating, appraising, and using contemporaneous research findings as the basis for clinical decisions"} \citep{rosenberg95}. A more concrete and widely accepted definition of EBM was coined by \citet{sacket96} who explained it as \emph{``the conscientious, explicit, and judicious use of current best evidence in making decisions about the care of individual patients}." As the practice has evolved, medical practitioners increasingly integrate, as required by the latest clinical guidelines, the latest clinical evidence with their own expertise. In addition, practitioners are also required to incorporate the informed patients' choices. The long-term objective of EBM is to enhance the standard of healthcare, promote practices that are proven by evidence to work, and to identify elements of the practice that do not work. A patient-oriented approach is key to the practice--- combining the best patient-oriented evidence with patient-centred care, placing the evidence in perspective with the needs and desires of the patient \citep{slawson05}. In the following sections, we provide an overview of EBM, discuss the problems associated with EBM practice, the resources and tools available for text processing in this domain and the solutions that NLP can offer.

Incorporation of research evidence when making clinical decisions involves \emph{``the practice of assessing the current problem in the light of the aggregated results of hundreds or thousands of comparable cases in a distant population sample, expressed in the language of probability and risk''} \citep{greenhalgh99}. EBM therefore involves the efficient use of information search strategies to locate reliable and up-to-date information from varying sources and extraction strategies to efficiently collect and analyse retrieved information. The practice includes a process formally known as the \emph{Critical Appraisal Exercise} which involves the following steps according to \citet{selvaraj10}:

\begin{enumerate}
\item clearly defining what the problem of a patient is, and what evidence is needed to address the patient's problems;
\item performing searches for the relevant literature efficiently;
\item selecting the best of the performed studies, and applying the guidelines of EBM to ascertain the validity of the studies;
\item appraising the quality of the evidence effectively; and
\item extracting and synthesising relevant evidences and applying them to the problem at hand.
\end{enumerate}

\emph{Defining a Patient Problem as a Clinical Question.}
Formulating the patient problem forms the basis for the clinical question, which is used to search resources for an evidence-based answer. A well formulated question includes information about a patient (symptoms, signs, test results and knowledge of previous treatments), the particular values and preferences of the patient, and other factors that could be relevant \citep{greenhalgh06}. All that information should be summarised into a succinct question defining the problem and the specific additional items of information needed to solve the problem. 


There has been substantial research in the area of medical question formulation and query-focused summarisation, and, in recent years, particularly in the field of EBM (\emph{e.g.}, a recent example of research drive in this area is the CLEF eHealth shared tasks \citep{suominen13}). This is because it has been shown that the answerability of questions can be largely increased by better query formulation among other things \citep{gorman95}. The PICO format, which has four components, has become the accepted framework for formulating patient-specific clinical queries \citep{richardson95}. The four components are: ``primary \textbf{P}roblem/\textbf{P}opulation, main \textbf{I}ntervention, main intervention \textbf{C}omparison, and \textbf{O}utcome of intervention." The PICO components of queries represent key aspects of patient care, and have become very much the standard for formulating EBM queries. Although this query formulation framework was originally designed for treatment-like queries, it was later extended to other kinds of medical queries \citep{armstrong99}. Research has shown that this mechanism of framing questions helps in the better formulating of clinical problems in queries, resulting in more accurate information retrieval results \citep{booth00,cheng04}. Although the PICO format has gained popularity with time, it is well known that not all clinical questions, because of the complex nature of the information needs, can be mapped in terms of PICO elements, and this is particularly true for non-therapy questions \citep{huang06}. There is also evidence that even doctors find it difficult to formulate the questions in terms of the PICO format \citep{ely02}. Variants of the framework have been suggested (\emph{e.g.}, PESICO \citep{schlosser06}, PIBOSO \citep{kim11}); they offer more flexibility and comprehensiveness and have applications beyond query formulation.

\emph{Conducting Literature Search.}
The medical domain has a massive amount of available published literature, scattered over many databases (\emph{e.g.}, MEDLINE\footnote{\url{http://www.ncbi.nlm.nih.gov/pubmed}. Accessed on 31st January, 2015.} indexes over 24 million articles), and searching for relevant literature requires expertise in this area. Searching for appropriate literature can include searching from raw databases, databases with search filters, databases of pre-appraised articles, databases of synthesised articles and even personal contact with human sources. Some research has been carried out on strategies for retrieving high quality medical articles \citep{haynes94,hunt97,shonjania01,montori04}. Searching for the correct literature is a rather tedious task, and it is in fact one of the major problems associated with EBM practice \citep{ely05}.

\begin{figure} [ht]
\centering
\scalebox{0.35}
{\includegraphics{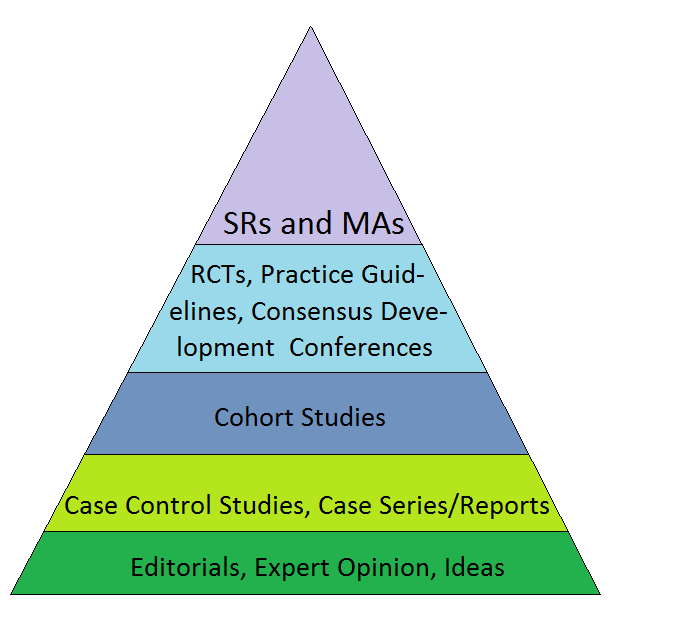}}
\caption{Quality of evidence with respect to publication types, adapted from \citet{gilbody96}.}
\label{fig1}
\end{figure}

\emph{Selecting the Best Resources.}
The selected articles must be closely relevant to the problem at hand, and, simultaneously, they must have a good \emph{level of evidence}. The level of evidence of a medical publication can be influenced by a number of factors such as the publication type and the sample size of the study. Checking the relevance of the papers requires a thorough analysis, and a ranking system is usually employed for examining the level of evidence of different sources. Medical publication types include Systematic Reviews (SR), Randomised Controlled Trials (RCT), Meta-Analyses (MA), single Case Studies, Tutorial Reviews and many more, including even personal opinions. Although all of them provide evidence of some form, their levels of evidence vary significantly. Figure \ref{fig1}, slightly modified from the one provided by \citet{gilbody96}, provides a ranking of some of the common medical article types (highest ranked on top). 

A detailed analysis of these studies is outside the scope of this survey. Guidelines are available for healthcare practitioners to obtain information from each of these types of studies and evaluate their levels of evidence. Examples of guidelines available online include ASSERT (A Standard for the Scientific and Ethical Review of Trials)\footnote{\url{http://www.assert-statement.org}. Accessed on  15th Jan, 2016.}, PRISMA (Preferred Reporting Items for Systematic Reviews and Meta-Analyses)\footnote{\url{http://prisma-statement.org}. Accessed on  15th Jan, 2016.}, and EQUATOR (Enhancing the QUAlity and Transparency Of health Research)\footnote{\url{http://www.equator-network.org}. Accessed on  15th Jan, 2016.}.

\emph{Identifying Strengths and Weaknesses.}
Once relevant papers are identified, each paper must be studied in detail to extract the evidence it contains with respect to the problem at hand. Practitioners are particularly interested in studying in detail what type of study was conducted, on the number of subjects, demographic information about the subjects,the kind of intervention, the length of the follow-up period, and the outcome measures \citep{greenhalgh06}. For quantitative tests, analysing the results of statistical tests is also an important task. Identification of the strengths and weaknesses provides a clear indication of the quality of the evidence provided by each article, and hence aids the practitioner to make the final decision.

\emph{Applying Evidence to Patient Care.}
Practitioners make their final judgment considering the outcomes presented in the article(s) and the relevance of the article(s) to the problem at hand. Often, a number of articles suggest the same solution or a similar one, making it easier for the practitioner to make a decision. However, there are also cases when chosen articles provide contradictory outcomes. In such cases, practitioners have to choose between outcomes based on personal experience, the \emph{quality of evidence} of the articles, the closeness of the articles to the given problem or other sources of evidence. Note that the quality of evidence  here refers to the overall quality of the evidence obtained from all the articles combined, whereas the level of evidence of a single article refers to the reliability of that particular article only. 

Generally, countries have mandatory clinical practice guidelines that must be adhered to when performing EBM. Practitioners primarily navigate through these guidelines during practice. As such, it is actually often the providers/developers of clinical guidelines who are required to perform elaborate literature searches and follow the approaches mentioned above for the preparation of the appropriate practice guidelines. 


There has been research to identify a grading system that is suitable for the practice of EBM, and numerous requirements for the systems have been specified, some more important than others in typical EBM settings. \citet{atkins04}, for example, suggested two aspects--- simplicity and clarity--- so that identifying the quality of a body of evidence is fast and can be easily projected on to some grading scale. \citet{ebell04} mentioned the comprehensiveness of evidence appraisal systems as a key aspect so that the same grading system may be applied to different types of clinical studies such as treatment, diagnosis and prognosis. 

\subsection{Resources for Evidence-based Medicine and the Medical Domain}
EBM requires practitioners to stay up-to-date with the latest medical literature and use the latest research discoveries. Practitioners currently utilise a variety of sources to obtain information. 

\subsubsection{The MEDLINE Database and Similar Databases}
The MEDLINE database is managed by the National Library of Medicine (NLM), U.S.A., and it is the most popular source of up-to-date evidence \citep{taylor03}. It indexes over 24 million articles covering a wide range of topics. It also provides various specialised indexing information and search techniques to aid retrieval. Despite the broad coverage of MEDLINE and its impressive collection, there are still journals that are not indexed by this database, particularly journals not published in the United States. Hence practitioners often have to refer to other similar databases that specialise in the required areas. The following is a list of databases that are relevant to various areas. Note that providing details of each database is beyond the scope of this survey, and therefore appropriate links to the databases are provided as footnotes. This list of databases is by no means exhaustive.

\begin{itemize}
\item Embase\footnote{ (\url{http://store.elsevier.com/embase}. Accessed on  15th Jan, 2016.)} -- This database is a popular resource for EBM, and covers over 8,500 biomedical journals. It indexes over 29 million articles, including all that are covered by MEDLINE. It is versatile and updated fast.
\item Allied and Complementary Medicine Database (AMED)\footnote{\url{https://www.ebscohost.com/academic/amed-the-allied-and-complementary-medicine-database}. Accessed on  15th Jan, 2016.} -- This is an alternative medicine database that is designed for practitioners and researchers interested in learning about alternative treatments. Topics include chiropractic, acupuncture, and so on, and many of the journals indexed are not available from other databases.
\item CINAHL\footnote{\url{http://www.cinahl.com}. Accessed on  15th Jan, 2016.} -- This is a dedicated database for nursing and allied health professionals.
Nursing, health education, social services in health care and other related disciplines are covered by this database.

\end{itemize}

\subsubsection{Databases of Synthesised Information}
One problem with MEDLINE and other `raw' databases is that they contain articles of varying quality -- from high quality SRs and MAs to informal, unreliable clinical trials. When specific topics are searched for in these databases, the results returned invariably contain a mixture of high and low quality articles. To address this problem, there are databases that only contain articles that are of high quality. A good example of such a database with synthesised evidence is the Cochrane Library.\footnote{\url{http://www.cochrane.org}. Accessed on  15th Jan, 2016.} It contains peer-reviewed Cochrane Systematic Reviews, Systematic Reviews listed in the Database of Abstracts of REviews (DARE), and selected published clinical trials in their Central Registry of Controlled Trials. However, the number of articles contained in this library is minute compared to the total number of medical articles available, and the scope of topics covered is understandably limited. Despite this, the Cochrane Library is very much the first port of call for clinical researchers looking for quality articles. 

Among such high quality databases, there are also guideline databases such as the National Guideline Clearing House\footnote{\url{http://www.guideline.gov}. Accessed on  15th Jan, 2016.} and National Institute for Health and Clinical Excellence (NICE)\footnote{\url{www.nice.org.uk}. Accessed on  15th Jan, 2016.} that provide evidence-based clinical practice guidelines. Many countries have their own national clinical practice guidelines, and sometimes their use are mandated by law (\emph{e.g.}, in Finland). There are also sources that not only synthesise the best available information but also present them in readily usable formats, such as: Clinical Evidence (CE),\footnote{\url{http://www.clinicalevidence.com/ceweb/conditions/index.jsp}. Accessed on  15th Jan, 2016.} Evidence-Based On-Call (EBOC).\footnote{\url{http://www.eboncall.org} Accessed on  15th Jan, 2016.} ACP Journal Club,\footnote{\url{https://acpjc.acponline.org/}. Accessed on 15th Jan, 2016.} Journal of Family Practice, \footnote{\url{http://www.jfponline.com/}. Accessed on 15th Jan, 2016.} Evidence Based Medicine,\footnote{\url{http://ebm.bmj.com/}. Accessed on 15th Jan, 2016.} and UpToDate\footnote{\url{http://uptodate.com/home/about/index.html}. Accessed on  15th Jan, 2016.}


There are several search engines specialised for the medical domain, and \url{about.com}\footnote{\url{http://websearch.about.com/od/enginesanddirectories/tp/medical.htm}. Accessed on  15th Jan, 2016.} provides an analysis of the top five search engines in the medical domain. The list includes PubMed, OmniMedicalSearch,\footnote{\url{http://www.omnimedicalsearch.com}. Accessed on  15th Jan, 2016.} WebMd,\footnote{\url{http://www.webmd.com}. Accessed on  15th Jan, 2016.} Healthline\footnote{\url{http://www.healthline.com}. Accessed on  15th Jan, 2016.} and HealthFinder.\footnote{\url{http://www.healthfinder.gov}. Accessed on  15th Jan, 2016.} Generic search engines are also frequently used to search for evidence and they, particularly Google, have evolved into effective search tools for online full text peer reviewed journals \citep{greenhalgh06}. According to a study by \cite{tutos10} that assessed the abilities of search engines to identify the correct information for meeting the information needs of clinical queries, Google performs better than any other systems including PubMed which is specialised on medical text. This clearly demonstrates the strength of generic search engines and their emergence as a useful tool in medical article retrieval and EBM practice.


\subsection{Obstacles to the Practice of EBM}
 
\citet{ely02} conducted a well known research to identify and explain the obstacles practitioners face when they attempt to answer clinical questions using evidence. The study revealed fifty-nine obstacles which were divided into the following five broad categories (i) recognising that there is a void in the knowledge of a practitioner on a specific topic, 
(ii) formulating a clinical query that accurately and comprehensively encapsulates the information needs of the practitioner, 
(iii) performing efficient searches for the relevant information, 
(iv) synthesising multiple and incomplete evidences to formulate the final response, 
(v) and using the formulated answer to make decisions about patient care, taking into account other relevant information such as the medical history of the patient. 
From the fifty nine obstacles, \citet{ely02} also identified a few that are considered to be critical by practising doctors, which are as follows:
\begin{enumerate}
\item The amount of time needed to find information is commonly seen as the most crucial obstacle in EBM practice. Most busy doctors lack the time or skills to track down and evaluate evidence, and when searching for evidence, practicing doctors do not have time to search multiple resources. Literature search and appraisal may take hours and even days. According to \citet{hersh02}, it takes more than 30 minutes on average for a practitioner to search for an answer. But usually practitioners spend about 2 minutes \citep{ely00}. Hence, many questions go unanswered. \citet{ely05} classify this problem as a `resource-related' problem and state that physicians want rapid access to concise answers that are easy to find and tell them what to do in specific terms. 

\item It is often difficult to structure a question that includes all the important information and is not vague. Recent research in medical information retrieval has focused on query formulation and other aspects of information retrieval to aid practitioners \citep{heppin12,kelly14}.

\item It is difficult to select an optimal strategy to search for information as the medical literature is \emph{unwieldy, disorganised and biased} \citep{godlee99}, and although electronic databases facilitate searching, the procedure requires skill and expertise. Even for an experienced librarian, the search to find a comprehensive set of documents related to a clinical query may take hours. Other difficulties related to information seeking include the presence of large numbers of irrelevant material in search results, and the difficulty in finding correct search terms \citep{verhoeven00}. 

\item An identified resource may not cover the specific  topic, and often practitioners may not have timely access to the necessary resources required to answer a question \citep{young01}. 
\item It is often difficult to tell when all the relevant evidence associated with a topic has been found. The number of relevant documents can vary significantly between topics, and, therefore, physicians are often left unsure as to when they should stop searching \citep{ely02,green05}. 

\item The synthesis of information--- once all the relevant information has been found, it is a daunting task to synthesise the information from different sources. Studies carried out have shown that practitioners frequently mention difficulties with generalising research findings and applying evidence to individual patient care \citep{young01}. The reasons behind this include the incapability of any of the sources to completely answer the clinical question, the selected articles not directly answering the clinical question, and different articles providing contradictory information \citep{ely05,ely02}.

\end{enumerate}

\subsection{Possible Applications of IR and NLP in EBM}
Despite the presence of many barriers, EBM practice has gained popularity over recent years for a number of reasons, including its promise of improving patient healthcare in the long run. As for the barriers, advances in technology are gradually eliminating them and making the process more efficient. From the problems and barriers specified earlier in this section, it can be inferred that the next boost in EBM practice will come from research in NLP. 

NLP offers suitable solutions to the problems faced by EBM practitioners, particularly for problems associated with information overload. Ideally, practitioners require bottom-line answers to their queries, along with estimates about the qualities of the evidences. They require fast access to information, and evidence-backed reasoning for recommendations \citep{ely05}. NLP has the potential of addressing all these requirements. For example, \emph{Query Analysis} can be used to understand and expand practitioners' queries through the use of domain-specific semantic information. Queries posed by practitioners are often very short with an average of 2.5 words \citep{hoogendam08}, and therefore existing techniques for ontology-driven query reformulation \citep{schwitter10} can be built upon to help practitioners compose and refine queries.  \emph{Information Retrieval} techniques tailored for the medical domain can be used to increase the recall and precision of literature searches. Strength of recommendation values can be used to classify documents to make searches more reliable. There has already been some research in this area with promising outcomes \citep{karimi09,pohl10}. \emph{Information Extraction} techniques which incorporate domain knowledge (\emph{e.g.}, MeSH terms, UMLS) can be used to extract relevant information, based on practitioners' questions, from the retrieved documents. This is also an area that is being explored by medical informatics researchers, and knowledge-based and statistical techniques have produced promising results \citep{demner07}. Furthermore, some research has attempted to extract semantic information and use that information to represent the main concepts presented in documents \citep{fiszman03,fiszman04,fiszman09}. Research in the area of information extraction is closely related to that of \emph{Document Summarisation}. The goal of summarisation in this context is to summarise the content extracted from multiple documents and present them to the users (\emph{i.e.}, specific bottom-line recommendations). Summarisation of multiple documents is a well explored area (\emph{e.g.}, in the news domain), however its application to the medical domain is still quite limited. Successful summarisation of medical documents and effective presentation of summarised information to the user are key to the successful development of end-to-end question answering systems that can be used for EBM practice. More specifically, query-focused multi-document summarisation that incorporates medical domain knowledge is an area of research that is worth exploring and success in this area will significantly advance the practice of EBM. For the rest of this survey, we cover automated summarisation techniques and their applications to the medical domain.


\section{Automated Text Summarisation}
In this section, we review the basics of automated summarisation techniques, important domain-independent approaches, breakthroughs in this research area, and approaches that have the potential of being applied to the medical domain. Since the field of ATS is too broad to be discussed in a single review, we have attempted to keep this section short with references to detailed literature for the interested reader.

\subsection{Overview}
According to \citet{radev02}, the intent of a summary is to express the informative contents of a document in a compressed manner. \citet{mani01} provides a more formal definition and explains that the process of summarisation involves ``\emph{taking an information source, extracting content from it, and presenting the most important content to the user in a condensed form and in a manner sensitive to the user's or application's needs}." \citet{sparck99} explains that summarisation is a hard task because it requires the characterisation of a source text as a whole, ``\emph{capturing its important content, where content is a matter of both information and its expression, and importance is a matter of what is essential as well as what is salient}." 

The motivation for building automated summarisation systems has increased over time due to the increasing availability of web-based textual information, and the explosion of available information has necessitated intensive research in this area \citep{mani00}. The number of online text documents has been increasing exponentially over the recent years, and this is also true for the medical domain \citep{sarker16summ}. Consequently, significant progress in automated summarisation has been made in the last two decades \citep{sparck07}. 

The prime advantage of having a large amount of available information is the redundancy in it. Research has shown how summarisation systems can exploit this redundancy \citep{barzilay05,clarke01,dumais02}, particularly when summarising from multiple documents. As will be discussed later, many summarisation systems rely on redundancy (\emph{e.g.}, frequently occurring words, concepts, etc.) to generate automated summaries. In EBM, redundancy is beneficial as it gives stronger evidence on a given finding. As for the cons, the abundance of information also introduces the need for efficient information retrieval and extraction, both of which are difficult tasks. In many cases, identification of relevant information requires elaborate manual searching through redundant information, which is often quite time consuming and therefore rather inefficient \citep{barzilay05}. ``\emph{It has been realised that added value is not gained merely through larger quantities of data, but through easier access to the required information at the right time and in the most suitable form}" \citep{afantenos05}. This gives rise to the need for technologies that can gather required information for the users and present them in a simplified, concise and friendly manner. Thus, we can say that while the large amount of information is a necessary condition for the development of automated summarisation systems, it also introduces the problem of information overload.   

\subsection{Factors of Text Summarisation}
Automated text summarisers must take into account a number of factors to achieve their goals. Here we discuss some of these factors, knowledge of which is essential to understand the process of text summarisation. The factors affecting ATS can be grouped into three main categories: input, purpose and output. We primarily draw the following information from existing literature  \citep{sparck99,mani01,afantenos05,sparck07}, and provide references to articles containing more detail in specific cases.

\subsubsection{Input}
The following factors are associated with the inputs of a summarisation system:

\emph{Unit} - Summarisers can take as input either one document or multiple documents. Summarising multiple documents represents	 a more difficult task than summarising single documents, and, additional algorithms are required to overcome problems of redundancy (different documents might present identical information which could lead to repetitions in the summary), inconsistency (the information presented in distinct documents may not be consistent), and incoherence (information extracted from separate documents may not be coherent when presented together). 

\emph{Language} - A summarising system can be mono-lingual, multi-lingual or cross-lingual. 

\emph{Domain} - Summarisation systems can either be domain-specific or domain-independent. Domain-independent approaches are applicable to documents from various domains without requiring any changes in the algorithm. While having the benefit of portability to different domains, domain-independent approaches fail to take advantage of knowledge and resources available to specific domains. Domain-specific systems are designed for specific domains and use all the available resources and knowledge available for the relevant domain. There has been significant research in text summarisation specific to domains such as news \citep{barzilay05,mckeown02}, scientific \citep{plaza11a}, legal \citep{farzinder04}, and medical \citep{mckeown98,demner07,rindflesch05}.


\emph{Structure} - This refers to the external structure of documents, such as headings, boxes, tables, and also more subtle exploitable structural information such as rhetorical patterns. The structure of documents can vary between domains. For example, news stories do not generally have much explicit structure other than top-level headlines \citep{sparck07}, while technical articles contain more structure which can be exploited by summarisation systems \citep{elhadad01,mckeown98,saggion02,demner07}. Medical article abstracts may be structured or unstructured, and this factor plays a vital role in the performance of summarisers.

\emph{Meta-data} - In some cases, header information or meta-data associated with input documents play an important role in summarisation. For example, dates of publications are crucial for news items. In the case of the medical domain, databases such as MEDLINE use meta-data to indicate the key topics in each article.


\subsubsection{Purpose}
These factors are associated with the purpose of the summaries. In other words, these factors determine what a summary is for or what a summary is like.


\emph{Summary Type} - Summaries can either be extractive or abstractive. Extractive summarisation involves extracting content from the source document(s) and presenting them as the summary. Abstractive summarisation, in contrast, involves discovering the most salient contents in the source document(s), aggregating them, and presenting them in a concise manner (\emph{e.g.} via techniques that automatically generate natural language) \citep{afantenos05}.

\emph{Information} - Summaries can be generic or user-oriented. Generic summaries only take into account the information found in the input document(s), whereas user-oriented summaries attempt to extract and summarise only the information that are needed to fulfill information requirements of the users. The summarisation query enables the user to formulate the information needs. 

\emph{Use} - This is the most important purpose factor \citep{sparck07}, and has a major influence on summary content and presentation. 

\emph{Audience} - Summaries must take into account the intended audience. Different summaries can be generated from the same input sources to suit the requirements of the audience. News material, for example, ``\emph{has at least two audiences: ordinary readers and analysts}" \citep{sparck07}. 
  
\emph{Time} - Time is a critical factor for some summarisation systems. For example, query-oriented summaries need to be delivered promptly, and thus, corresponding systems need to be very fast. 


\subsubsection{Output} These factors are associated with the summarised output.

\emph{Coverage} - Coverage of a set of sources by a summary can  be either \emph{comprehensive or selective} \citep{sparck07}. \emph{Reflective summaries are comprehensive} (\emph{i.e.}, summarises the whole source text), while \emph{query- or topic-oriented summaries are selective} (\emph{i.e.}, summarises only a segment of the source text that contains information associated with the query/topic) \citep{sparck07}.

\emph{Compression} - This refers to the amount by which the information from the source(s) are reduced during summarisation.

\emph{Structure} - The output of a summary may be plain text or the information may be represented using tables (\emph{e.g.}, \citet{mani00}), forms (\citet{farzinder04}), and lists (\emph{e.g.}, \citet{radev00}) to name a few. 



\subsection{Approaches to Summarisation}

\subsubsection{Early Summarisation Approaches}
The earliest known work on automated summarisation dates back to 1958 when \citet{luhn58} proposed that the \emph{frequency} of words could provide a measure of their importance in a document. In his work, he ranked words based on their frequencies and used the ranking of individual  words in a sentence to calculate their \emph{significance}, finally selecting the top ranked sentences as the summary. \citet{baxendale58} used \emph{sentence position} as a feature for selecting important sentences in a document (\emph{e.g.}, for news sentences, early sentences are more important than later sentences). \citet{edmundson69} extended this work on summarisation and proposed a  extensible model for extractive summarisation, which later on came to be known as the \emph{Edmundsonian Paradigm} \citep{mani01}. In this proposed model, the author used a weighted linear equation to combine four features to score text nuggets (\emph{e.g.}, sentences) for summary inclusion:
\begin{equation}
W(s) = \alpha C(s) + \beta K(s) + \gamma L(s) + \delta T(s)
\end{equation}
where  $\alpha , \beta , \gamma$  and  $\delta$ are manually assigned weights; \emph{W} is the overall weight of sentence $s$, \emph{C} represents the score given to sentence $s$ due to the presence of cue words (bonus words or stigma words) extracted from a corpus, \emph{K} represents the score given for key words (based on word frequency), \emph{L} is the score given based on sentence location features, and \emph{T} is the weight assigned based on terms in the sentence that are also present in the title. Such a statistical approach to sentence selection has been very popular in the research community with \citet{earl70} being the first to experiment with a variety of shallow lexical features. More recently, numerous statistical approaches using multiple words, noun phrases, main verbs, and named entities. have been proposed \citep{barzilay97,lin00,harabagiu05,hovy99,lacatusu03}, while some research focused on utilising shallow lexical items with salient properties (\emph{e.g.}, subheadings) \citep{teufel97,chakrabarti01} or via the analysis of topical content \citep{ando00}. The importance of the discourse structure of a text was realised quite early in summarisation research, and this property has been heavily exploited ever since \citep{hearst94,marcu98,hahn97}.
\subsubsection{Recent Advances in Automatic Summarisation}
Following the early summarisation research works, research on summary generation and evaluation techniques have been boosted by the Document Understanding Conference (DUC) and similar other regular workshops and initiatives. They include: the NII Test Collection for Information Retrieval Project (NTCIR)\footnote{\url{http://research.nii.ac.jp/ntcir/index-en.html}. Accessed on  15th Jan, 2016.}, the Text REtrieval Conference (TREC)\footnote{\url{http://trec.nist.gov/}. Accessed on  15th Jan, 2016.}, and the Text Analysis Conference (TAC)\footnote{\url{http://www.nist.gov/tac/}. Accessed on  15th Jan, 2016.}, which was initiated from the Text Summarisation track of the DUC and the Question Answering track of the TREC. Since the commencement of widespread research on ATS, a branch of research has focused on the application of machine learning algorithms for text summarisation. In most of the initial research, machine learning was employed sparingly, as preliminary steps \citep{sparck07}. Early research mostly assumed feature independence and used the \emph{Na\"ive Bayes} classifier \citep{kupiec95,lin97,aone99} with various features including those used in the \emph{Edmundsonian Paradigm}. Later research introduced the use of richer feature sets and a range of machine learning algorithms. For example, \citet{lin99} used \emph{Decision Trees}, \citet{conroy01} employed \emph{Hidden Markov Models}, \citet{osborne02} applied a \emph{Log Linear Model} for sentence extraction, \citet{svore07} used \emph{Neural Nets}, and \citet{schilder08} used \emph{Support Vector Machines}. Interested readers can find more information about these approaches and other important related work in \citet{sparck07}.
 
While one branch of research focused on machine learning approaches, another branch progressed research on natural language analysis techniques. \citet{miike94} and \citet{marcu98,marcu99,marcu00}  use the Rhetorical Structure Theory (RST)\footnote{See \url{http://www.di.uniba.it/intint/people/fior\_file/INTINT05/RST.pdf}. Accessed on  15th Jan, 2016.} by building RST source text trees using discourse structure information, and identifying the nuclei of the trees to build the summaries. \citet{polanyi04} and \citet{thione04} utilise the PALSUMM model which uses more abstract discourse trees. \citet{barzilay97} show the use of \emph{lexical chains} --- sequences of related words that can span short or long distances --- for single-document summarisation. For example, \citet{mckeown98} and \citet{elhadad01} use a template suited to the medical domain; \citet{mckeown95a} and \citet{radev98} fill up template slots from a database of news information as a first step to their algorithm for news summarisation; and \citet{sauper09} show the use of templates to automatically generate Wikipedia articles.

\subsubsection{Multi-document Summarisation}
The field of multi-document summarisation was pioneered by the SUMMONS \citep{mckeown95a,radev98} summarisation system, which employed a template-based summarisation approach. Extractive summarisation systems have been shown to work well for multiple documents particularly in the news domain, an example being the MEAD\footnote{Available at: \url{http://www. summarisation.com/mead/}. Accessed on  15th Jan, 2016.} \citep{radev00a} system. Multi-document summarisation approaches suffer from the problems of incoherence and redundancy, and numerous approaches have been proposed to address them. One popular approach to reduce redundancy is the use of \emph{clustering}. In this technique, common themes or concepts across document sets are identified and grouped or clustered together. Once the clusters are created, the summary can be generated by applying various algorithms that depend primarily on the content and compression needs. For example, some use a single sentence to represent each cluster in the final summary \citep{mckeown95a,radev00,yih07}, while some generate a composite sentence from each cluster through the use of information fusion so as to combine the most salient concepts from multiple sentences within a cluster \citep{barzilay97,barzilay99,barzilay05}. 

Another approach that has been successfully applied to limit the selection redundant information when performing summarisation is Maximal Marginal Relevance (MMR). This technique, which is specifically useful for query-focused summarisation, text segments that are deemed relevant to a query, usually via some text similarity measure, are rewarded. At the same time, redundant text segments are penalised using the same similarity measure. The two similarity measures are combined using a linear equation in an attempt to balance redundancy and relevance. MMR was initially utilised for document retrieval and is given by the following formula:

\begin{equation}
MMR \equiv arg max_{D_{x}\in R\backslash S}[\lambda (Sim_{1}(D_{x},Q) - (1-\lambda) max_{D_{y}\in S} Sim_{2}(D_{x},D_{y}))]
\label{mmr}
\end{equation}
where, as explained by \citet{carbonell98}: ``$Q$ is a query; $R = IR(C, Q, \theta )$, i.e., the ranked list of documents retrieved by an IR system given a document collection $C$ and a relevance threshold $\theta$; $S$ is the subset of documents in $R$ already selected; $R \backslash S$ is the set difference, i.e., the set of unselected documents in $R$; $Sim_{1}$ is the similarity metric used in document retrieval and relevance ranking between documents and a query; and $Sim_{2}$ can be the same or a different metric."

Graph-based approaches have also been applied to text summarisation \citep{mani97,mani99,erkan04,leskovec05}, with \citet{mani97} being the pioneers in this area. In their approach, the authors use nodes to represent words  and edges between nodes represent relationships. The summaries generated can be topic driven, and there is no text in the summaries. Instead, the summary content is represented via as nodes and edges that represent contents and relations between them. When summarising a pair of documents, \emph{common nodes} represent same words or synonyms, while \emph{difference nodes} are those that are not common. Sentence selection from the graph is computed from the average activated weights of the covered words: for a sentence $s$, its score in terms of coverage of common nodes is given by the following formula:
\begin{equation}
score(s) = \frac{1}{|c(s)|}\sum_{i=1}^{|c(s)|} weight(w_{i})
\end{equation}
where $c(s) = \{w|w\in Common \bigcap s\}$.
The score for \emph{differences} is calculated similarly. The sentences with higher \emph{common} and \emph{difference} scores are selected for the final summary. 

\citet{erkan04} presented the LexRank system for multi-document summarisation, which is a graph-based system using a connected, undirected graph to represent documents. A similar method, suitable for single-document summarisation only, was proposed by \citet{mihalcea11}. Other graph based approaches have been proposed, both in the medical domain \citep{reeve07,fiszman04} and outside it \citep{litvak08}.

Similar to graph-based techniques are centroid-based summarisation techniques, first proposed by \citet{radev00a,radev04}. The summarisation is done in three stages. The first stage involves grouping together news articles on the same topics (topic detection), with the goal of clustering together news stories based on the events they discuss. These groups are called clusters and the \emph{centroid} for each cluster is computed. The centroid of a cluster is a pseudo-document consisting of a set of highly relevant words. The relevance of these words is determined using Term Frequency-Inverse Document Frequency (\emph{tf$\times$idf}) measures---those having tf$\times$idf values above a predetermined threshold are included in the centroid. Tf and idf values are computed using the following equations: 

\begin{equation}
tf_{i,j} = \frac{n_{i,j}}{\sum_{k}n_{kj}} ; idf_{i} = \log \frac{|D|}{|\{d:t_{i}\in d\}|}
\end{equation} 
where $n_{i,j}$ is the occurrence frequency of the considered term $t_i$ in document $d_j$, and $\sum_{k}n_{kj}$ is the sum of the number of occurrences of all the terms in $d_j$; $|$D$|$ is the total number of documents the corpus contains, and $|\{d:t_{i}\in d\}|$ is the total count of documents where the term $t_i$ occurs. In the second stage, salient topics are identified via the use of centroids, and the overall score for each sentence is computed by combining three distinct values--- centroid, positional, and overlap with the first sentence--- minus a penalty for redundancy.

\section{Summarisation and Question Answering for the Medical Domain}
In this section, we provide an overview of text summarisation and Question Answering (QA) approaches targeted towards the medical domain. The goal here is to present some of the important research work in this area and analytically review more recent and promising approaches. The medical domain itself is quite broad, and so we attempt to adhere to research work that is relevant for EBM. It must be mentioned that, in the recent past, there has been steady ongoing research in biomedical and medical text  processing \citep{afantenos05,zweigenbaum07}. However, compared to other domains, there is very little published research in summarisation and question answering \citep{fiszman09}. A variety of tools and resources have also been made available to aid summarisation approaches in this domain.

\subsection{NLP Tools for Summarisation}
Summarisation of text in the medical domain requires the incorporation of significant amounts of domain-specific knowledge. There are numerous tools and knowledge resources currently available. Here we briefly introduce some of the important ones, starting with the resources.

\subsubsection{UMLS}
The Unified Medical Language System (UMLS)\footnote{\url{http://www.nlm.nih.gov/index.html}} is a system and resource developed by the U.S. National Library of Medicine (NLM) for biomedical vocabularies. The development and maintainance of this system is targeted to aid language processing systems in the biomedical domain. The goal is to create standards and mappings for terminologies, and relationships between terminologies that automated systems can understand. More specifically, it was developed as ``\emph{an effort to overcome two significant barriers to the effective retrieval of machine-readable information}: the variety of names used to express the same concept and the absence of a standard format for distributing terminologies"  \citep{lindberg93}. The UMLS consists of the following three major components:
\begin{itemize}
\item Metathesaurus -- This is the major component of the UMLS and consists of a repository of inter-related biomedical concepts and terms obtained from several controlled vocabularies and their relationships. 
\item Semantic Network -- This provides a set of high-level categories and relationships used to categorise and relate the entries in the Metathesaurus. Each concept in the Metathesaurus is given a `semantic type' and certain `semantic relationships' may be present between members of the various semantic types. For example, a \emph{disease} mention (\emph{e.g.}, headache) may have an \emph{is-treated-by} relationship with a \emph{drug} mention (\emph{e.g.,} aspirin).
\item SPECIALIST Lexicon -- This is a database of lexicographic information for use in NLP. Each entry in it contains syntactic, morphological and orthographic (spelling) information.

\end{itemize}
\subsubsection{SNOMED CT}
SNOMED CT\footnote{\url{http://www.ihtsdo.org}. Accessed on  15th Jan, 2016.} is the most comprehensive source of medical terminology and is a standard  in the U.S. for the exchange of health information electronically. It can be accessed via the NLM and the National Cancer Institute (NCI). It is one of the controlled vocabularies used by the UMLS.
\subsubsection{MetaMap}
The MetaMap\footnote{\url{http://metamap.nlm.nih.gov}. Accessed on 14th May, 2014.} is a software that can map biomedical text to UMLS Metathesaurus, which was developed and is managed by the NLM. It also employs word sense disambiguation and a part-of-speech tagger that is designed specifically for biomedical text. 

\subsubsection{SemRep}
SemRep\footnote{\url{http://semrep.nlm.nih.gov}. Accessed on 14th May, 2014.} is a programalso developed at the NLM. It identifies \emph{semantic predications} ( \emph{i.e.}, subject-relation-object triples) in biomedical natural language. It has been used for a variety of biomedical applications, including automated summarisation, literature-based discovery and hypothesis generation.
\subsubsection{Other Tools}
There are other text processing tools that have been used in the past for medical text processing although their functionalities are not restricted to this domain. The following is a brief list:
\begin{itemize}
\item GATE\footnote{\url{http://gate.ac.uk}. Accessed on 17th May, 2014.}: It is a Java platform from the University of Sheffield, and, according to their website, it is capable of solving almost any text processing problem.
\item LingPipe\footnote{\url{http://alias-i.com/lingpipe}. Accessed on 18th May, 2014.}: A useful, Java-based language processing tool that is free for non-commercial use and is widely used for research.
\item OpenNLP\footnote{\url{http://opennlp.sourceforge.net}. Accessed on 25th May, 2014.}: It is a source for a variety of Java-based NLP tools that can perform a range of basic and advanced text processing tasks.
\end{itemize}
\subsection{Overview of the Medical Domain and Decision Support Systems}
A number of factors make the medical domain a complex and interesting one for text processing. They include: large volume of data (\emph{e.g.}, about 24 million articles in MEDLINE alone); highly complex domain-specific terminologies (\emph{e.g.}, drug names and disease names); domain-specific linguistic and ontological resources (such as UMLS); and software tools and methods for identifying semantic concepts and relationships (such as MetaMap \citep{aronson01}). 

The integration of technology with medical practice was initiated through the use of Clinical Decision Support (CDS) systems. Such systems help practitioners ``\emph{make clinical decisions, deal with medical data or with the knowledge of medicine necessary to interpret such data}" \citep{shortliffe90}. Early CDS systems consisted mostly of applications that facilitated diagnosis, treatment \citep{shortliffe79,miller82} and the management of patients (\emph{e.g.}, through computerised guidelines and alerts) \citep{barnett83}. However, the capabilities of such applications were quite limited, primarily because they did not have access to raw medical data \citep{friedman99}. 

Since a lot of medical data is textual, the need to integrate NLP with CDS systems became more noticeable as the volume of medical data increased. This requirement introduced new challenges as well, such as faster processing \citep{demner09}, which the latest research in large-scale distributed processing has successfully addressed. A detailed discussion of CDS systems is outside the scope of this paper and in the following subsections, we focus instead on a subset of CDS systems that attempt to provide answers to clinical queries by summarising the information contained in medical texts. Such systems consist of information extraction, summarisation, and QA components, and we do not distinguish between these three types. Identifying and presenting evidence in a condensed manner is essentially a task of summarisation. Hence we refer to these systems/components as summarisation systems. For QA systems, we primarily discuss their summarisation components. For the interested reader, \citet{friedman99} provide a review of early NLP-based CDS approaches and a recent and detailed analysis of CDS systems and NLP for the medical domain is provided by \cite{demner09}. Readers unfamiliar with QA may refer to \cite{molla07} for an overview of QA techniques in restricted domains and \cite{athenikos09} for a review of QA approaches in the biomedical domain.

\subsection{Overview of Medical Text Summarisation Approaches}
\subsubsection{Information Extraction Approaches}
Early summarisation systems were mostly concerned with extracting relevant information from structured or unstructured medical text. The Linguistic String Project (LSP) \citep{sager94} is an example of early medical NLP work. Its primary purpose is the transformation of clinical narratives into formal representations. MedLEE \citep{friedman05} is also responsible for extracting information from clinical narratives and presenting the information in structured forms through the use of a controlled vocabulary. It is used for processing various forms of notes and reports and is integrated with a clinical information system. HiTEx \citep{zeng06} is yet another system used for extracting findings such as diagnoses and family history from clinical narratives through the use of NLP techniques. TRESTLE (Text Retrieval Extraction and summarisation Technologies for Large Enterprises) \citep{gaizauskas01} is an information extraction system that generates summarised information from pharmaceutical newsletters in one sentence through the identification of Named Entities (NEs) followed by sentence extraction based on the presence of key NEs. Drug and disease names are considered by this system to be named entities. \citet{hahn02} present MEDSYNDIKATE, a natural language processor that automatically extracts clinical information from reports. The contents of the texts are transferred to conceptual representations that correspond to a knowledge base, and the system incorporates domain knowledge to semantically interpret major syntactic patterns in medical documents. \citet{xu10} propose the MedEx system which uses NLP to extract medication information from clinical notes with very high recall and precision. 

\emph{Extractive Summarisation Approaches}. Most summarisation systems in this domain applied extractive summarisation approaches like the initial MiTAP system (MITRE Text and Audio Processing) \citep{damianos02} which was targeted towards the monitoring of infectious disease outbreaks or other biological threats. MiTAP monitors various sources of information such as online news, television news, and newswire feeds. and captures information which are filtered and processed to identify sentences, paragraphs, and POS tags. The final summary is generated by WebSumm \citep{mani99} as extracted sentences from the processed text. \citet{reeve07} propose a single-document, extractive summarisation approach that combines BioChain \citep{reeve06a}, which identifies relevant sentences using concept-chaining (similar to lexical chaining but applied to UMLS concepts), and the FreqDist system \citep{reeve06b}, which uses a frequency distribution model to identify relevant sentences. This hybrid approach, called ChainFreq, sequentially employs the BioChain method to identify candidate sentences that contain relevant concepts, and then the FreqDist method, which generates a set of summary sentences from the chosen sentences. Other extractive approaches have been proposed with various intents: \citet{mihalcea04} presents an approach for automated sentence extraction using graph based ranking algorithms; \citet{elhadad06} proposes extractive algorithms for performing user-sensitive text summarisation; and, more recently, \citet{abacha11} put forth some summarisation approaches for the \emph{automated extraction of semantic relations between medical entities}.

\emph{Non-extractive Approaches}. MUSI is an early system that applies semantic information thoroughly \citep{lenci02} (MUltilingual summarisation for the Internet). It follows an approach similar to the ones already mentioned for sentence extraction but also has the capability of producing semantic representations of the extracted sentences to produce an abstractive summary.  Other than the MUSI system, early research in this area did not consider abstractive summarisation. Recently, however, a number of abstractive summarisation systems have been proposed (\emph{e.g.}, PERSIVAL \citep{elhadad01}, MedQA \citep{lee06a,yu07a}, EpoCare \citep{niu05,niu06}. We discuss some of these approaches in more detail in the following subsections. 

\subsubsection{Progress in Medical Summarisation and Question Answering}
In recent years, medical document summarisation has received significant research attention, which attempt to make use of greater computational power and the availability of more sources of domain knowledge. A variety of document types are now addressed by the different approaches, and QA systems have been developed specific to this domain. Most work on query-focused summarisation in this research area has been executed under the broader domain of QA. While early NLP focused on domain-independent QA, restricted domain QA has received focus over the last decade \citep{molla07}. Restricted domain QA, particularly in the medical domain, require incorporation of terminological and specialised domain-specific knowledge \cite{zweigenbaum03,zweigenbaum09}.

Some research has focused primarily on question analysis and the coarse-grained classification as the first step for medical QA \citep{athenikos09}. \citet{yu05a}, \citet{yu05b,yu07a} and \citet{yu08}, for example, study the answerability of clinical questions and attempt to classify clinical questions based on the Evidence Taxonomy \citep{ely99}, and also into general topics. The QA approach proposed by \citet{weiming07} relies heavily on semantic information present on the questions and documents. First, candidate sentences are identified and phrases are extracted by identifying mappings between question and answer semantic types. The system has been evaluated for factoid and complex questions and is shown to have very good recall and precision (77\% and 92\% respectively). The use of medical semantic types to identify salient information in sentences based on query needs have been exploited in later, more comprehensive summarisation systems, which we discuss in the next subsection. Finally, \citet{workman12} present a dynamic summarisation approach using an algorithm called \emph{Combo} to identify salient semantic predications. It was shown to outperform several baseline methods in terms of recall and precision.
 

\subsection{Detailed Review of Systems: The EBM Perspective}
While discussing approaches in the previous subsection, we attempted to only provide the reader with a flavour of the various directions in which medical text summarisation has progressed. We intentionally skipped some important systems and filtered out many. In this subsection, we discuss the characteristics of a number of systems that perform summarisation of medical text. For full QA systems, we primarily focus on their summarisation components. This review not only discusses the capabilities of the systems but also presents their strengths and weaknesses from the perspective of EBM. Table \ref{font-table} provides a summary comparison of the systems mentioned here.\footnote{Note that in the table, SemRep represents the summarisation approach using SemRep proposed by \cite{fiszman09}.}

\begin{landscape}
\begin{table} [t]

\begin{tabular}{@{}|l|p{3.1cm} p{3.5cm} p{3.0cm} p{3.5cm} p{2.5cm}|@{}} \toprule 

\textbf{System} & \textbf{Input} & \textbf{Unit}  & \textbf{Use of Semantic Information} &\textbf{Summary Type} & \textbf{Target user} \\ 
\midrule \midrule 

MedQA* & MEDLINE and the Web & Multi-document& No & Non-extractive/Definitional & Healthcare Practitioners\\
   
CQA 1.0* & MEDLINE abstracts & Multi-document but separate summaries for each abstract & Yes (UMLS) & Extractive & Healthcare Practitioners \\ 


SemRep & Clinical trials from MEDLINE & Multi-document & Yes (UMLS and SemRep)  & Non-extractive & Healthcare Practitioners\\ 

EpoCare & MEDLINE abstracts (cited by CE articles) & Multi-document & Yes (UMLS) & Extractive & Healthcare Practitioners\\   

PERSIVAL & Patient Records, Medical Articles and Web-based Text Articles & Multi-document & Yes (UMLS) & Non-extractive & Healthcare Practitioners and Laymen\\ 

AskHermes* & MEDLINE abstracts and full texts & Multi-document & Yes (UMLS) & Extractive &  Healthcare Practitioners \\ 

QSpec & MEDLINE abstracts & Single- and multi-document & Yes (UMLS) & Extractive & Healthcare Practitioners\\

\bottomrule

\end{tabular} 

\caption{\label{font-table} Comparison of summarisation systems for the medical domain. Systems available online are marked with a *.}
\end{table}

\end{landscape}

\subsubsection{MedQA}
MedQA\footnote{Available at: \url{http://askhermes.org/MedQA/}. Accessed on  15th Jan, 2016.} \citep{lee06a,yu07a} answers definitional questions by producing paragraph-level answers from MEDLINE and the web. Syntactic parsing, query formulation, and query classification techniques are used to prepare queries, and an IR engine retrieves relevant documents \citep{yu05a,yu05b}. The answer extraction component employs document \emph{zone detection} and sentence categorisation via the identification of cue phrases. Definitional sentences are identified in this fashion.  Hierarchical clustering \citep{lee06b} and centroid-based summarisation techniques \citep{radev00} are used for text summarisation. The system's ability to answer definitional questions is evaluated against three search engines--- Google, OneLook and PubMed \citet{yu07a,yu07b} Evaluations show that Google is very effective in obtaining definitions, outperforming MedQA.  

MedQA has opened new directions in medical text summarisation. It has shown how supervised classification can be used for intermediate steps in medical text summarisation, such as query analysis. MedQA, however, is not capable of incorporating semantic information, and the intent of the system is very different from the requirements of the EBM practitioner. The key limitation of the system is that it can only answer definitional questions, and not real life-like complex ones. 

\subsubsection{CQA 1.0}
\cite{demner07} present a QA system\footnote{Available at: \url{http://www.umiacs.umd.edu/~demner/}. Accessed on  15th Jan, 2016.} that uses a statistical and knowledge-based approach and is particularly targeted towards the practice of EBM. The proposed system uses PICO representations of questions as queries which are sent to PubMed to retrieve an initial set of abstracts. From the abstracts, each of the PICO elements (Problem/Population, Intervention, Comparison and Outcome) are extracted using various techniques. MetaMap is extensively used to identify UMLS terms and their categories. The population extractor uses a series of hand-crafted rules to identify occurrences of population terms in the abstracts, with preference given to terms occurring earlier in the documents. The problem extractor identifies elements that represent the UMLS semantic group `DISORDER'. The intervention and comparison elements are identified in a similar way--- by recognition of multiple UMLS semantic types (\emph{e.g.},  \emph{clinical drug}, \emph{diagnostic procedure}, etc.). In structured abstracts, more weight is given if the semantic types occur in `title', `aims' or `methods' sections, while, in unstructured abstracts, more weight is given if they appear towards the beginning of the document. Using the extracted knowledge, the authors re-rank the retrieved documents using a document ranking algorithm that takes into account the knowledge elements, strength of evidence and other task specific considerations \citep{lin06}. An outcome extractor extracts outcome sentences from the retrieved documents using a ``\emph{strategy based on an ensemble of classifiers (a rule-based classifier, a bag-of-words classifier, an n-gram classifier, a position classifier, an abstract length classifier and a semantic classifier)}" \citep{lin06}. Each sentence is given a probability based on the classifier scores, and the top-ranked sentences are chosen. As the final output, the system simply produces the top ranked sentences from the top re-ranked documents along with the question and the strength of recommendation. Only basic clinical questions such as `\emph{What is the best drug treatment for X?}' are addressed. The authors evaluate the performance of the knowledge extractors against different baselines and also manually evaluate the final output against a baseline that only presents top sentences from unranked documents. 

The techniques applied by the CQA system show the importance of statistics and domain-specific knowledge for medical text summarisation. There are several limitations of the system. It relies on PICO frames for queries (instead of natural language questions), and an information synthesis technique is absent at the end to produce a single answer from related documents. Furthermore, the algorithm applied to predict the qualities of evidences does not follow an evidence-based guideline. 

\subsubsection{Summarisation using SemRep}
\citet{fiszman04}, \citet{rindflesch05} and \citet{fiszman09} propose an approach to abstractive summarisation that primarily relies on utilising semantic information. The summariser uses SemRep as a semantic processor to perform source interpretation and predication listing, and relies on user-specified topics. A transformation stage generalises and condenses the list of predicates generating a conceptual condensate for the input topic. 
The transformation is carried out in four phases: Relevance, Connectivity, Novelty, and Saliency. First, the relevance phase is responsible for identifying predications for a particular topic (\emph{e.g.}, treatments). The UMLS semantic network is utilised, along with a controlled schema, to identify predications. The predications, which must conform to the schema, are called the `\emph{core predications}. The Connectivity phase is a generalisation process and retrieves all predications that share arguments with core predications. The Novelty phase condenses the \emph{summaries} further by removing general argument predications (\emph{e.g.}, Pharmaceutical Preparations), which are predications that appear higher in the UMLS metathesaurus hierarchy.  Finally, the Saliency phase calculates the occurrence of uses simple frequency measures to keep the representative predicates, predications, and arguments. The final summary is produced in the form of a graph, and the approach can be applied to both single documents and multiple documents without requiring any modifications. 

One of the drawbacks of the system is that it does not incorporate query information. The system is evaluated \citep{fiszman09} on its capability to discover drug interventions for disorders/syndromes only (other forms of interventions are not taken into account). Therefore, the application domain of the system is very limited. Furthermore, only clinical trials are used as source documents.  The performance of the system is compared to a baseline that selects drug names based on the frequencies of their occurrence in source texts. A scoring mechanism called the \emph{clinical usefulness score} is used. It rewards the systems for identifying beneficial drug interventions and penalises them for identifying harmful or not useful ones.  To assess the usefulness of drug interventions, drug listings in \emph{Clinical Evidence} (CE) articles are used as gold standards, in place of manually annotated references. The system is shown to outperform the baseline in terms of both the clinical usefulness score and mean average precision (MAP).  

The summarisation approach applied by the SemRep system is simple, innovative, and effective. Its performance illustrates the importance of domain-specific semantic information, and the usefulness of distributional semantics in automated summarisation for this domain. Incorporating query-focus into the summarisation procedure could technique more applicable for EBM practice. Summary information requirements must be identified from clinical questions instead of manually identified topics, and the summarisation component should be able to identify information other than drug interventions. 

\subsubsection{EPoCare}
The EpoCare (Evidence at Point of Care)\footnote{\url{http://www.cs.toronto.edu/km/epocare/index.html}. Accessed on  15th Jan, 2016.} project  \citep{niu03,niu04,niu05,niu06} is an initiative by the University of Toronto to develop a clinical QA system. The current implementation relies heavily on automatic identification of PICO elements from both clinical questions and their corresponding answers. PICO keywords are first identified from the question and used as keywords for retrieval. The problem of identifying answers to a clinical question is divided into four sub-problems -- (i) identifying roles (PICO elements) in the text, (ii) identifying the  lexical boundary of each element, (iii) analysing the relationships between distinct elements and (iv) determining which combinations of roles are most likely to contain correct answers. The initial work presented by Niu and colleagues addresses simple treatment-type questions. MetaMap is used for automated identification of interventions and problems. The authors note that identification of outcomes is a much more difficult task, and they identify cue words (nouns, verbs and adjectives) that indicate the presence of outcomes in sentences. The outcome detection task is further subdivided into two sub-tasks --- outcome identification and lexical boundary determination. In addition to sentence level outcome detection, the authors suggest that the polarity of outcomes play an important role in determining which sentences to choose as answers. Four categories of polarities are defined (positive, negative, no outcome and neutral). The authors use Support Vector Machines (SVMs) to classify sentences into the four categories, and show that best results are obtained by combining linguistic features with domain features.

The outcomes of the polarity classification task are used in a multi-document summarisation approach to automatically find information from MEDLINE abstracts to answer a clinical question. Presence and polarity of outcomes, position of sentence in abstracts, length of sentences and Maximal Marginal Relevance (MMR) are used as features in a machine learning algorithm (SVM) used to solve the summarisation problem. Sentences in MEDLINE abstracts cited by Clinical Evidence (CE) articles are manually annotated for each clinical query, and sentences obtained from the automated approach are compared with these for evaluation. A total of 197 abstracts from 24 subsections in CE are annotated to give a total of 2,298 sentences. 

The outcome detection task is shown to have an accuracy of 83\%. The polarity assessment task is shown to have an accuracy of 79.42\%. For the summarisation task, it is observed that the identification of outcomes and polarity improves performance significantly. However, F-scores are shown to be very low (\textless 0.50) in all cases. ROUGE is also used for evaluation but shows little variation in performance for distinct combinations of features. The outcomes of the EpoCare project show that abstractive summarisation approaches, which utilise automatic polarity classification of sentences, have the potential to be applied for the generation of evidence-based summaries. 

\subsubsection{PERSIVAL}
PERSIVAL (PErsonalized Retrieval and summarisation of Image, Video and Language) \citep{elhadad01} is a medical digital library designed to provide customised \emph{access to a distributed library of multimedia medical literature}. It is not possible to discuss the whole project in this review, and hence we focus on the text summarisation component of the system that produces customised, abstractive summaries for persons from technical and non-technical backgrounds \citep{elhadad05,elhadad06}. 

The summarisation system takes as input three different sources: patient records, medical articles (about cardiology) that are appropriate for the patient, and a query by the user from which key words are extracted. The input articles are classified as prognosis, treatment or diagnosis. Relevant sentences from the `Results' sections of the articles are extracted and stored in in a pre-defined template, which contains three sets of information: parameter(s), findings, and relation. The relation describes the relation between a parameter and a finding. The information extraction phase also identifies the extent to which the parameters are dependent, along with other meta-data, such as the position of the sentence from which a parameter has been extracted \citep{elhadad05}. Following the extraction, the information is filtered to keep only the portions that are relevant to the patient's medical records, and hand-crafted templates are filled with the extracted elements.  In the next step, the filled templates are converted into semantic representations via mapping them into a graph. Repetitions (when two nodes are connected via multiple vertices of the same type) and contradictions (when two nodes are connected by multiple vertices of different types) are identified from the graphs, and this information is used to generate a coherent summary. The representation is then ordered for summary generation: relations that are deemed relevant to the user question are given the highest preference followed by recitations and contradictions, which in turn are followed by preference based on the relation type (\emph{e.g.}, relations representing risks are given higher priority than association relations). Finally, dependent relations identified from one template are output together, and the final summary is generated using natural language generation techniques, along with hyperlinks to medical concepts. 

The summaries generated by the system are not evidence-based, and the intent of the system is to provide personalised information for users from both technical and non-technical backgrounds. The approaches applied by this system have the potential of being applied for evidence-based summarisation.

\subsubsection{AskHERMES}
AskHERMES\footnote{Available at: \url{http://www.askhermes.org/index2.html}. Accessed on  15th Jan, 2016.} \citep{cao11} is a clinical QA system that is capable of taking queries in natural language, performs thorough query analysis, and generates single-document, query-focused extracts from a group of texts that are relevant to the query. The input to the system, being in natural language, requires minimal query formulation by the user, and the interface allows quick navigation of the summaries. The system is demonstrated to outperform Google and UpToDate when answering complex clinical questions.

AskHermes operates in five phases: a \emph{query analysis} module automatically identifies the information needs posed by questions by generating a list of informative query terms; a \emph{related questions extraction} module identifies a list of questions that are similar, based on the terms; an \emph{information retrieval} module returns the relevant documents; an \emph{information extraction} module identifies relevant passages from the source documents; and a \emph{summarisation and answer presentation} module that analyses the identified relevant passages, identifies and removes redundant elements, and finally outputs summaries with structure.

In the query analysis phase, a query is first classified into one of 12 general topics \citep{yu08}, and then \emph{keywords} are automatically extracted from the question using Conditional Random Fields. Following the retrieval of relevant documents, a two-layer hierarchical clustering is applied to group passages into different topics. Query terms and their UMLS mappings are used to assign the cluster labels. The first layer of clustering generates the topic labels for a tree structures, and a second layer of clustering is applied to provide more refined categories. The final set of clusters are ranked based on the \emph{key} query terms that are found in in them, and redundancies are detected and removed using \emph{longest common substrings}.

AskHERMES is shown to perform comparably to state-of-the-art systems, and its potential application in the field of EBM is very promising. One issue is that the lengthy summaries do not satisfy the requirement of bottom-line recommendations required by practitioners. Due to the extractive nature of the summarisation, it is difficult to synthesise information from multiple documents and generate brief summaries. However, the system's performance suggests that customising the summary generation process to the type of question may be beneficial. Furthermore, the authors show that key query terms can be used to determine topics. 

\subsubsection{QSpec}
QSpec\footnote{\url{http://web.science.mq.edu.au/~diego/medicalnlp/}. Accessed on 2nd February, 2015.} \cite{sarker14thesis,sarker12alta,sarker13aime,sarker15aiim,sarker16summ} is a summarisation system that attempts to generate evidence-based summaries to complex medical queries posed by practitioners.The system utilises a publicly available data set\footnote{\url{http://sourceforge.net/projects/ebmsumcorpus/}. Accessed on 2nd February, 2015.} \cite{molla15}, which consists of real-life clinical queries posed by practitioners, evidence-based justifications in response to the queries (single-document summaries), and bottom-line recommendations (multi-document summaries). The proposed system consists of two primary components: (i) query-focused summariser \cite{sarker16summ}, and (ii) automatic evidence grader \cite{sarker15aiim}. The summariser module performs single-document, extractive summarisation based on multiple domain-specific and domain-independent features, such as sentence positions, lengths, sentence classes, UMLS semantic types, and others. The system takes as input questions in natural language. Following the retrieval of the relevant documents,\footnote{The system does not perform information retrieval.} single-document summaries are generated by selecting the three most appropriate sentences from each document. The crucial part of this component is the sentence selection process, which utilises statistics derived from the annotated data in the corpus. Using statistics associated with a range of topics (\emph{e.g.}, question-specific semantic types, associations between semantic types, maximal marginal relevance and others), the system attempts to apply weights to each sentence score, and finally ranks the sentences based on the scores.

Multi-document summaries are generated by performing automatic, contextual polarity classification of sentences \cite{sarker13ijcnlp}. In addition to the multi-document summaries for each evidence, a supervised learning approach that utilises a sequence of classifiers \cite{sarker14thesis} is applied to grade the quality of the evidence.

The system is shown to perform significantly better than several baseline and benchmark systems on a scale that employs ROUGE F-scores for relative comparison of systems \cite{ceylan10}. However, the generation of multi-document summaries is described as a much more difficult task, and the contextual polarity classifier is only applied to a subset of the corpus that addresses therapeutic questions. This system illustrates the usefulness of annotated corpora for this task, and it is likely that future research will utilise the publicly available data set that this system has been trained on.

\section{Evaluation}
The objective of this section is to briefly discuss the evaluation techniques used for automated summarisation, approaches attempted in the past and possible approaches for evaluating summaries for EBM. Evaluation of automatically generated summaries is a hard problem \citep{sparck99,sparck07}. This is primarily because of the fact that it is a heuristic problem -- there are more than one acceptable solution, and a universally accepted standard evaluation is absent. Evaluation measures usually focus on specific features of the summarised text, which depend largely on the summarisation factors. The features focused upon by evaluation techniques for generic summaries can be significantly different from those for query-focused summaries. Evaluation techniques can also vary according to the unit of summarisation (i.e., single \emph{vs.} multi-document), domain, type (extractive \emph{vs.} non-extractive) and other factors mentioned previously. In this brief section, we focus only on approaches that have been used or may be relevant for application in EBM summarisers. The interested reader can refer to \cite{lin02} for a brief discussion on automatic and manual evaluation techniques. 

\subsection{Extrinsic and Intrinsic Evaluation}
Techniques for automatic summary evaluation can be divided into two broad classes--- extrinsic and intrinsic. Extrinsic evaluation techniques focus on the purpose of summaries. The objective is to measure the usefulness of a summary for a specific task. The single feature that plays the most important role in determining the usefulness of a summary is its content, and numerous evaluation techniques use this feature as an indication of the qualities of summaries. Recall, precision, F-score, coverage and other similar measures \citep{lin02} have been frequently used in the past as simple evaluations of content relevance (\emph{e.g.}, \cite{weiming07,niu03,niu04,niu05,niu06}). More sophisticated techniques have also been applied for extrinsic evaluation. An example is the assessment of the answerability of questions, which can be answered from original texts, from summarised texts \citet{morris92,teufel01,mani02}. These evaluation techniques have been employed in open domain QA, but have not been explored for the medical domain until recently \cite{sarker12}. Such an evaluation approach can be useful for summarisers targeted towards EBM, as discussed later.  

Intrinsic methods concentrate on the summary itself, trying to measure features such as coherence, cohesion, grammaticality, readability and other important features. Intrinsic methods for extractive summaries assess features such as discourse well-formedness, while those for non-extractive approaches must also assess sentence well-formedness. Although in some domains, such as news, intrinsic evaluation plays an important role, for a query-focused summarisation system for EBM, we believe it to be much less important. The reasons are discussed later in the section.

\subsection{Evaluation Techniques}
Both manual and automated evaluation techniques are used for evaluating summaries. The following are some common techniques used in both methods of evaluation. 

\subsubsection{Gold Standards}
Gold standards (also known as human reference summaries) are often used for evaluating automatic summarisation primarily because humans can be ``\emph{relied upon to capture important source content and to produce well-formed output text}" \citep{sparck07,suominen08}. The expected output summaries are manually created by human experts, who are often experts in a specific domain. The created summaries therefore contain the necessary content and become the target performance for systems. Evaluation then compares the generated summaries with the gold standard summaries. This can be done automatically or manually, and is itself a research problem. The more similar a generated summary is to the gold standard, the better it is considered to be. 

For years, the absence of gold standard data has been an obstacle to summarisation research in the medical domain. However, the recent creation of specialised corpora for EBM \cite{molla15} is driving the development of data-driven summarisation and evaluation techniques. 

\subsubsection{Baselines}
Baselines can be considered to be the opposite of gold standards in that they indicate the minimum level of required performance by a summarisation system. For extractive summarisation, various baselines such as \emph{n} random sentences have been used. A more appropriate baseline for news summarisation was introduced by \citet{brandow95} who used the first \emph{n} sentences. With these baselines, the minimum performance required by a system is to select \emph{n} sentences that better summarise the source than the baseline. Similar baselines have been established for summarisation in various other domains. For example, \citet{demner07} propose an outcome extractor that is compared against a baseline of `last \emph{n} sentences' (since outcomes presented in a medical paper usually appear towards the end of the abstract). Baseline measures of tf.idf type have also been used in the literature and even for EBM \citep{fiszman09}. In such baselines, the summarisation units (sentences, words, n-grams) are chosen based on their tf.idf values. A standard baseline for summarisation systems across domains and even within specific domains, however, still does not exist.

\subsubsection{Topic-oriented Evaluation}
Topic-oriented evaluation techniques are specialised to the topic and intent of the summarisation task. A number of topic-oriented evaluation schemes have been proposed in the literature, both within the medical domain and outside. A recent example of a topic-oriented evaluation mechanism is the `Clinical Usefulness Score' (CUS) \citep{fiszman09}, a unique evaluation of generated summaries. The CUS is a categorical performance metric. In calculating this score, interventions extracted by a system are assigned to one of four high-level categories depending on how they match the interventions in a predetermined reference standard. The goal is to give credit to a system for finding beneficial interventions and, similarly, penalise it for finding harmful interventions. 



\subsection{Manual Evaluation}
Due to the difficulty associated with evaluating automatic summaries, manual evaluation is still a common practice. There is more confidence in manual evaluation (compared to automatic evaluation) since humans can infer, paraphrase and use world knowledge to relate to text units with similar meanings but worded differently. In such evaluations, domain experts (often several for a single summary) read and grade summaries, usually using some chosen scale. Most of the systems mentioned in the last subsection of the previous section have undergone some form of manual evaluation, or at least involved human experts for the preparation of a gold standard (\emph{e.g.}, \citep{demner07}). However, agreement among human summarisers is generally quite low, and the process of manual evaluation is quite expensive. Human judgements have been shown to be unstable and inconsistent as well \citep{lin02}. As a result, alternative automatic evaluations having high correlation with human scores are usually used. 

\citet{nenkova04} present the pyramid approach for the manual evaluation of summaries. Summary content is divided into summarisation content units (SCU), and SCUs representing the same semantic information are manually annotated in each source document. Once annotation is complete, all the SCUs are assigned weights based on the number of summaries each of them appear in. Next, the SCUs are partitioned into a pyramid in which each tier contains SCUs of the same weight and higher tiers contain SCUs of higher score (i.e., SCUs appearing in more human summaries). Summaries that contain more top-tier SCUs are ranked higher than those that contain lower tier SCUs. An optimal summary consists of all the top-tier SCUs combined, within the summary length limits. The score for an automatically generated summary is calculated by summing the weights the constituent SCUs, and dividing this number by the sum of the SCUs of the corresponding optimal summary.



An earlier approach presented by \citet{halteren03} is similar: atomic semantic units, called factoids, are used to represent the meanings of sentences. The approach requires the generation of a large number of summarised articles from which the gold standard can be obtained by identifying the most frequently occurring factoids. As an example, the authors show that to generate a news article summary of a 100 words (from 50 sample summaries), all factoids appearing in at least 30\% of the summaries had to be included. Hence, gold standards of different lengths can be generated by varying the factoid threshold. Although this approach ensures very high agreement among the humans, its requirement of a large number of sample summaries makes it quite an expensive approach.  

\subsection{Automatic Evaluation}

\subsubsection{ROUGE}
Recall-Oriented Understudy for Gisting Evaluation (ROUGE) \citep{lin03,lin04} is one of the most widely used tools for automatic summary evaluation, and consists of a number of metrics that evaluate summaries based on distinct criteria. The various ROUGE measures attempt to compute similarities between system generated summaries and gold standard summaries at a lexical level. One of the metrics, for example, called ROUGE-N, is an n-gram based recall-oriented measure, and is calculated using the following formula:

\begin{equation}
ROUGE-N (s) = \frac{\sum_{r\in R} <\phi_{n} (r), \phi_{n}(s)>}{\sum_{r\in R} <\phi_{n} (r), \phi_{n}(r)>} 
\end{equation}
where $R = \{r_{1},...,r_{m}\}$ denote the group of gold-standard summaries, $s$ denotes an automated system summary, and $\phi_{n}(d)$ denotes a binary vector contained in a document $d$ \citep{lin03,lin04}. This metric, therefore, simply attempts to measure the extent to which an automatically generated summary contains the same information as the reference summary. Other ROUGE metrics apply different techniques with the same primary intent. For example, ROUGE-L attempts to find the \emph{longest common subsequence} (LCS) between two summaries, with the rationale that summaries with longer LCSs are more similar. Another ROUGE metric, known as ROUGE-S, is a \emph{gappy} alternative of the ROUGE-N metric (for n = 2), and matches ordered bi-grams of the generated summary with reference summaries.


One problem with using ROUGE for evaluation is that it gives an absolute value score for a generated summary. As a result, it is difficult to ascertain how good or bad a summarisation system's score is compared to the best possible obtainable score. It is also difficult to perform relative comparisons between systems. To address these drawbacks, \citet{ceylan10} paved the way for relative comparisons between ROUGE scores using a percentile-rank based approach. \citet{ceylan10} empirically showed that since the number of extractive summaries for a set of documents are finite, the ROUGE scores for summary sentence combinations fall within a finite range, following a gaussian distribution. Most combinations get a ROUGE score that is very close to the mean. This leads to a long-tailed probability distribution for all ROUGE scores, meaning that a small increase in the ROUGE score assigned to a system can indicate large increases in percentile ranks \cite{sarker16summ}. Importantly, the distribution of all ROUGE scores for a particular data set can be used to perform relative comparisons between systems, and such a technique has been applied recently for comparative evaluation of systems.

\subsubsection{BLEU}
BLEU \citep{papineni02} was originally designed for evaluating machine translation, and has been shown to be promising for automatic summary evaluation as well \citep{lin02}. In this technique, \emph{automatically computed accumulative n-gram matching scores (NAMS)} between a model unit (MU) and a system summary (S) are used as performance indicators of the system \citep{papineni02}. A number of combinations of n-grams are used to compute NAMS, and the technique is shown to have satisfactory correlation with human scores.

\subsubsection{Information-theoretic Evaluation}

\citet{lin06a} introduce an information-theoretic approach to the automatic evaluation of summaries based on the divergence of distribution of terms between an automatic summary and model summaries. For a set of documents \emph{D}, the authors assume that there exists a probabilistic distribution with parameters specified by $\theta_{R}$ that generates reference summaries from \emph{D}, and the task of summarisation is to estimate $\theta_{R}$. Similarly, the authors assume that every system summary is generated from a probabilistic distribution with parameters specified by $\theta_{A}$. The process of summary evaluation then becomes the task of estimating the distance between $\theta_{R}$ and $\theta_{A}$. The authors present a number of variants of divergence measures (\emph{e.g.}, Jensen-Shannon divergence (JS), Kullback-Leibler divergence (KL)) for this and show that this technique is comparable to ROUGE for the evaluation of single-document summarisation, and better than ROUGE in evaluating multi-document summarisation systems. \citet{louis08}  and \citet{kabadjov10} present similar information theoretic evaluation approaches, which incorporate other divergence measures and also attempt to capture complex phoenomena such as synonymy. 


\subsubsection{ParaEval}
The motivation behind this evaluation technique is the lack of semantic matching of content in automatic evaluation. The authors of this evaluation technique \citep{zhou06} explain that an essential part of semantic matching involves paraphrase matching and this evaluation system attempts to perform that. ParaEval applies a three-level comparison strategy. At the top level, an optimal search via dynamic programming to find multi-word to multi-word paraphrase matches between generated and reference summaries is used. In the second level, a greedy algorithm is used to find single-word paraphrase matches among non-matching fragments in the first level. Finally, literal lexical uni-gram matching is performed on the remaining text at the third level. The authors show that the quality of ParaEval's evaluations closely resembles that of ROUGE.

\subsection{Discussion of Evaluation Techniques for EBM}
If summarising for EBM is a hard task, evaluating summaries is even harder. A summarisation system for EBM should be capable of extracting evidence from medical articles and additionally assess the grade of the evidence. The evaluation should be able to determine if the  evidence is correctly extracted and also if the extracted information correctly answers the practitioner's query. Therefore, the single most important aspect of the summary is its content and a strong focus on extrinsic evaluation is required. Intrinsic evaluation to assess aspects such as coherence and readability are perhaps not very important, and a stronger focus needs to be on the purpose of summaries \cite{sparck99}.

For the automatic evaluation of summaries, approaches based on n-gram co-occurrence such as ROUGE are the most frequently used. Although these approaches are very robust and efficient, their main drawback is that if two summaries were produced using non- or almost non-overlapping vocabulary, yet conveying the same information, the similarity score such summaries would be assigned by purely n-gram based metrics would be too low and, hence, unrepresentative of the actual information they share.  This is definitely not desirable for evaluation, particularly in the medical domain where relations such as synonymy and hyponymy play important roles. Furthermore, making relative comparisons between different summarisers is difficult using automatic evaluation approaches such as ROUGE. Thus, approaches that incorporate domain knowledge, semantic similarities, and comparisons between summarisation systems are required. Research on the evaluation of summarisers for EBM is still very much in its infancy. However, recent works in this area such as that of \citet{fiszman09} and \citet{sarker16summ} have made useful contributions by ensuring that evaluation techniques assess the performance of the summariser in the light of its goals.

\section{Conclusions}
In this paper, we provided an overview of EBM and described how summarisation of text can aid practitioners at point-of-care. We discussed the obstacles that EBM practitioners face, as indicated by various research papers on the topic. Following our review of the domain, we provided an overview of automatic summarisation, its intent, and some important contributions to automatic summarisation research. We discussed that unlike automatic summarisation research on some domains such as news, the medical domain has not received much research attention. We also explained that domain-independent summarisation techniques lack sufficient domain knowledge, incorporation of which can be crucial for summarisation research. We reviewed several recent summarisation systems that are customised for the medical domain. Our review of these systems revealed promising approaches, including the clever use of domain-specific information and distributional semantics. Our survey indicates that combining some of the useful approaches from existing literature, and building on from these already explored techniques, may produce encouraging results for text summarisation in this domain. Our survey reveals that an important factor limiting summarisation research in this domain has been the lack of suitable corpora/gold standard summaries. However, recent trends in the creation of specialised corpora will inevitably drive the development of more data-centric systems in the future.



\bibliographystyle{ACM-Reference-Format-Journals}
\bibliography{phd_papers_update}


\begin{thebibliography}{00}


\ifx \showCODEN    \undefined \def \showCODEN     #1{\unskip}     \fi
\ifx \showDOI      \undefined \def \showDOI       #1{{\tt DOI:}\penalty0{#1}\ }
  \fi
\ifx \showISBNx    \undefined \def \showISBNx     #1{\unskip}     \fi
\ifx \showISBNxiii \undefined \def \showISBNxiii  #1{\unskip}     \fi
\ifx \showISSN     \undefined \def \showISSN      #1{\unskip}     \fi
\ifx \showLCCN     \undefined \def \showLCCN      #1{\unskip}     \fi
\ifx \shownote     \undefined \def \shownote      #1{#1}          \fi
\ifx \showarticletitle \undefined \def \showarticletitle #1{#1}   \fi
\ifx \showURL      \undefined \def \showURL       #1{#1}          \fi

\bibitem[\protect\citeauthoryear{Afantenos, Karkaletsis, and
  Stamatopoulos}{Afantenos et~al\mbox{.}}{2005}]%
        {afantenos05}
{S.~D. Afantenos}, {V. Karkaletsis}, {and} {P. Stamatopoulos}. 2005.
\newblock \showarticletitle{Summarization from Medical Documents: A Survey}.
\newblock {\em Artificial Intelligence in Medicine\/} {33}, 2 (2005), 157--177.
\newblock


\bibitem[\protect\citeauthoryear{Ando, Boguraev, Byrd, and Neff}{Ando
  et~al\mbox{.}}{2000}]%
        {ando00}
{R.~K. Ando}, {B.~K. Boguraev}, {R.~J. Byrd}, {and} {M.~S. Neff}. 2000.
\newblock \showarticletitle{{Multi-document Summarization by Visualizing
  Topical Content}}. In {\em Proceedings of the NAACL-ANLP Workshop on
  Automatic Summarization}. 79--88.
\newblock


\bibitem[\protect\citeauthoryear{Aone}{Aone}{1999}]%
        {aone99}
{O.~C. Aone}. 1999.
\newblock {\em Advances in Automatic Text Summarization}.
\newblock MIT Press, Chapter A Trainable Summarizer with Knowledge Acquired
  from Robust NLP Techniques, 71--80.
\newblock


\bibitem[\protect\citeauthoryear{Armstrong}{Armstrong}{1999}]%
        {armstrong99}
{E.~C. Armstrong}. 1999.
\newblock \showarticletitle{The well-built clinical question: the key to
  finding the best evidence efficiently.}
\newblock {\em Wisconsin Medical Journal\/} {98}, 2 (1999), 25--28.
\newblock


\bibitem[\protect\citeauthoryear{Aronson}{Aronson}{2001}]%
        {aronson01}
{A.~R. Aronson}. 2001.
\newblock \showarticletitle{{Effective Mapping of Biomedical Text to the UMLS
  Metathesaurus: The MetaMap Program}}. In {\em {Proceedings AMIA Annual
  Symposium}}. 17--21.
\newblock


\bibitem[\protect\citeauthoryear{Athenikos and Han}{Athenikos and Han}{2009}]%
        {athenikos09}
{S.~J. Athenikos} {and} {H. Han}. 2009.
\newblock \showarticletitle{{Biomedical question answering: A survey}}.
\newblock {\em Computer Methods and Programs in Biomedicine\/} {99}, 1 (2009),
  1--24.
\newblock


\bibitem[\protect\citeauthoryear{Atkins, Best, Briss, Eccles, Falck-Ytter,
  \emph{et al.}, and the G.~R. A. D. E. Working~Group}{Atkins
  et~al\mbox{.}}{2004}]%
        {atkins04}
{D. Atkins}, {D. Best}, {P.~A. Briss}, {M. Eccles}, {Y. Falck-Ytter},
  {S.~Flottorp \emph{et al.}}, {and} {the G.~R. A. D. E. Working~Group}. 2004.
\newblock \showarticletitle{Grading quality of evidence and strength of
  recommendations.}
\newblock {\em BMJ\/} {328}, 7454 (June 2004), 1490--1497.
\newblock


\bibitem[\protect\citeauthoryear{Barnett, Winickoff, Morgan, and
  Zielstorff}{Barnett et~al\mbox{.}}{1983}]%
        {barnett83}
{G.~O. Barnett}, {R.~N. Winickoff}, {M.~M. Morgan}, {and} {R.~D. Zielstorff}.
  1983.
\newblock \showarticletitle{{A Computer-Based Monitoring System for Follow-Up
  of Elevated Blood Pressure}}.
\newblock {\em Medical Care\/} {21}, 4 (1983), 400--409.
\newblock


\bibitem[\protect\citeauthoryear{Barzilay and Elhadad}{Barzilay and
  Elhadad}{1997}]%
        {barzilay97}
{R. Barzilay} {and} {M. Elhadad}. 1997.
\newblock \showarticletitle{Using Lexical Chains for Text Summarization}. In
  {\em Proceedings of the ACL Workshop on Intelligent Scalable Text
  Summarization}. 10--17.
\newblock


\bibitem[\protect\citeauthoryear{Barzilay and McKeown}{Barzilay and
  McKeown}{2005}]%
        {barzilay05}
{R. Barzilay} {and} {K. McKeown}. 2005.
\newblock \showarticletitle{{Sentence Fusion for Multidocument News
  Summarization}}.
\newblock {\em Computational Linguistics\/} {31}, 3 (2005), 297--328.
\newblock


\bibitem[\protect\citeauthoryear{Barzilay, McKeown, and Elhadad}{Barzilay
  et~al\mbox{.}}{1999}]%
        {barzilay99}
{R. Barzilay}, {K. McKeown}, {and} {M. Elhadad}. 1999.
\newblock \showarticletitle{Information Fusion in the Context of Multi-Document
  Summarization}. In {\em {Proceedings of the 37th annual meeting of ACL}}.
  550--557.
\newblock


\bibitem[\protect\citeauthoryear{Baxendale}{Baxendale}{1958}]%
        {baxendale58}
{P. Baxendale}. 1958.
\newblock \showarticletitle{Machine-made index for Technical Literature - An
  Experiment}.
\newblock {\em IBM Journal of Research Development\/} {2}, 4 (1958), 354--361.
\newblock


\bibitem[\protect\citeauthoryear{{Ben Abacha} and Zweigenbaum}{{Ben Abacha} and
  Zweigenbaum}{2011}]%
        {abacha11}
{A. {Ben Abacha}} {and} {P. Zweigenbaum}. 2011.
\newblock \showarticletitle{{Automatic extraction of semantic relations between
  medical entities: a rule based approach}}.
\newblock {\em {Journal of Biomedical Semantics}\/} {2}, 5 (2011), S4.
\newblock


\bibitem[\protect\citeauthoryear{Booth, O'Rourke, and Ford}{Booth
  et~al\mbox{.}}{2000}]%
        {booth00}
{A. Booth}, {A.~J. O'Rourke}, {and} {N.~J. Ford}. 2000.
\newblock \showarticletitle{Structuring the pre-search reference interview: a
  useful technique for handling clinical questions}.
\newblock {\em {Bulletin of the Medical Library Association}\/} {88}, 3 (July
  2000), 239--246.
\newblock


\bibitem[\protect\citeauthoryear{Brandow, Mitze, and Rau}{Brandow
  et~al\mbox{.}}{1995}]%
        {brandow95}
{R. Brandow}, {K. Mitze}, {and} {L.~F. Rau}. 1995.
\newblock \showarticletitle{Automatic condensation of electronic publications
  by sentence selection}.
\newblock {\em Information Processing and Management\/} {31}, 5 (1995),
  675--686.
\newblock


\bibitem[\protect\citeauthoryear{Cao, Liu, Simpson, Antieau, Bennett, Cimino,
  Ely, and Yu}{Cao et~al\mbox{.}}{2011}]%
        {cao11}
{Y. Cao}, {F. Liu}, {P. Simpson}, {L.~D. Antieau}, {A. Bennett}, {J.~J.
  Cimino}, {J.~W. Ely}, {and} {H. Yu}. 2011.
\newblock \showarticletitle{{AskHermes: An Online Question Answering System for
  Complex Clinical Querstions}}.
\newblock {\em Journal of Biomedical Informatics\/} {44}, 2 (2011), 277 -- 288.
\newblock


\bibitem[\protect\citeauthoryear{Carbonell and Goldstein}{Carbonell and
  Goldstein}{1998}]%
        {carbonell98}
{J. Carbonell} {and} {J. Goldstein}. 1998.
\newblock \showarticletitle{{The use of MMR, diversity-based reranking for
  reordering documents and producing summaries}}. In {\em Proceedings of the
  International ACM-SIGIR Conference on Research and Development in Information
  Retrieval}. 335--336.
\newblock


\bibitem[\protect\citeauthoryear{Ceylan, Mihalcea, \"{O}zertem, Lloret, and
  Palomar}{Ceylan et~al\mbox{.}}{2010}]%
        {ceylan10}
{H. Ceylan}, {R. Mihalcea}, {U. \"{O}zertem}, {E. Lloret}, {and} {M. Palomar}.
  2010.
\newblock \showarticletitle{Quantifying the Limits and Success of Extractive
  Summarization Systems Across Domains}. In {\em Proceedings of NAACL}.
  903--911.
\newblock


\bibitem[\protect\citeauthoryear{Chakrabarti, Joshi, and Tawde}{Chakrabarti
  et~al\mbox{.}}{2001}]%
        {chakrabarti01}
{S. Chakrabarti}, {M. Joshi}, {and} {V. Tawde}. 2001.
\newblock \showarticletitle{Enhanced Topic Distillation Using Text, Markup
  tags, and Hyperlinks}. In {\em Proceedings of the International ACM-SIGIR
  Conference on Research and Development in Information Retrieval}. 208--216.
\newblock


\bibitem[\protect\citeauthoryear{Cheng}{Cheng}{2004}]%
        {cheng04}
{G.~Y. Cheng}. 2004.
\newblock \showarticletitle{A study of clinical questions posed by hospital
  clinicians}.
\newblock {\em {Journal of the Medical Library Association}\/} {92}, 3 (2004),
  445--458.
\newblock


\bibitem[\protect\citeauthoryear{Clarke, Cormack, and Lynam}{Clarke
  et~al\mbox{.}}{2001}]%
        {clarke01}
{C.~L.~A. Clarke}, {G.~V. Cormack}, {and} {T.~R. Lynam}. 2001.
\newblock \showarticletitle{Exploiting redundancy in question answering}. In
  {\em Proceedings of the International ACM-SIGIR Conference on Research and
  Development in IR}. 358--365.
\newblock


\bibitem[\protect\citeauthoryear{Conroy and O'Leary}{Conroy and
  O'Leary}{2001}]%
        {conroy01}
{J.~M. Conroy} {and} {D.~P. O'Leary}. 2001.
\newblock \showarticletitle{{Text Summarization via Hidden Markov Models}}. In
  {\em Proceedings of the International ACM-SIGIR Conference on Research and
  Development in IR}. 406--407.
\newblock


\bibitem[\protect\citeauthoryear{Damianos, Wohlever, Ponte, Wilson, Reeder,
  McEntee, Kozierok, Hirschman, and Day}{Damianos et~al\mbox{.}}{2002}]%
        {damianos02}
{L. Damianos}, {S. Wohlever}, {J. Ponte}, {G. Wilson}, {F. Reeder}, {T.
  McEntee}, {R. Kozierok}, {L. Hirschman}, {and} {D. Day}. 2002.
\newblock \showarticletitle{{Real users, real data, real problems: the MiTAP
  system for monitoring bio events}}. In {\em {Proceedings of the second
  international conference on Human Language Technology Research}}. 357--362.
\newblock


\bibitem[\protect\citeauthoryear{{Demner-Fushman}, Chapman, and
  McDonald}{{Demner-Fushman} et~al\mbox{.}}{2009}]%
        {demner09}
{D. {Demner-Fushman}}, {W.~W. Chapman}, {and} {C.~J. McDonald}. 2009.
\newblock \showarticletitle{{Methodological Review: What can natural language
  processing do for clinical decision support?}}
\newblock {\em Journal of Biomedical Informatics\/} {42}, 5 (October 2009),
  760--772.
\newblock


\bibitem[\protect\citeauthoryear{Demner-Fushman and Lin}{Demner-Fushman and
  Lin}{2007}]%
        {demner07}
{D. Demner-Fushman} {and} {J.~J. Lin}. 2007.
\newblock \showarticletitle{Answering Clinical Questions with Knowledge-Based
  and Statistical Techniques}.
\newblock {\em Computational Linguistics\/} {33}, 1 (2007), 63--103.
\newblock


\bibitem[\protect\citeauthoryear{Dumais, Banko, Brill, Lin, and Ng}{Dumais
  et~al\mbox{.}}{2002}]%
        {dumais02}
{S. Dumais}, {M. Banko}, {E. Brill}, {J. Lin}, {and} {A. Ng}. 2002.
\newblock \showarticletitle{Web question answering: is more always better?}. In
  {\em Proceedings of the International ACM-SIGIR Conference on Research and
  Development in Information Retrieval}. 291--298.
\newblock


\bibitem[\protect\citeauthoryear{Earl}{Earl}{1970}]%
        {earl70}
{L. Earl}. 1970.
\newblock \showarticletitle{Experiments in Automatic Extracting and Indexing}.
\newblock {\em Information Storage and Retrieval\/}  {6} (1970), 313--334.
\newblock


\bibitem[\protect\citeauthoryear{Ebell, Siwek, Weiss, Woolf, Susman, Ewigman,
  and Bowman}{Ebell et~al\mbox{.}}{2004}]%
        {ebell04}
{M.~H. Ebell}, {J. Siwek}, {B.~D. Weiss}, {S.~H. Woolf}, {J. Susman}, {B.
  Ewigman}, {and} {M. Bowman}. 2004.
\newblock \showarticletitle{{Strength of recommendation taxonomy (SORT): a
  patient-centered approach to grading evidence in the medical literature}}.
\newblock {\em {American Family Physician}\/} {69}, 3 (Feb. 2004), 548--556.
\newblock


\bibitem[\protect\citeauthoryear{Edmundson}{Edmundson}{1969}]%
        {edmundson69}
{H.~P. Edmundson}. 1969.
\newblock \showarticletitle{New Methods in Automatic Extracting}.
\newblock {\em {Journal of the ACM}\/} {16}, 2 (1969), 264--285.
\newblock


\bibitem[\protect\citeauthoryear{Elhadad}{Elhadad}{2006}]%
        {elhadad06}
{N. Elhadad}. 2006.
\newblock {\em User-Sensitive Text Summarization: Application to the Medical
  Domain}.
\newblock Ph.D. Dissertation. Columbia University.
\newblock


\bibitem[\protect\citeauthoryear{Elhadad, Kan, Klavans, and McKeown}{Elhadad
  et~al\mbox{.}}{2005}]%
        {elhadad05}
{N. Elhadad}, {M. Kan}, {J.~L. Klavans}, {and} {K. McKeown}. 2005.
\newblock \showarticletitle{Customization in a Unified Framework for
  Summarizing Medical Literature}.
\newblock {\em {Artificial Intelligence in Medicine}\/} {33}, 2 (2005),
  179--198.
\newblock


\bibitem[\protect\citeauthoryear{Elhadad and McKeown}{Elhadad and
  McKeown}{2001}]%
        {elhadad01}
{N. Elhadad} {and} {K.~R. McKeown}. 2001.
\newblock \showarticletitle{Towards generating patient specific summaries of
  medical articles}. In {\em {Proceedings of the NAACL 2001 Workshop on
  Automatic Summarization}}. 31--39.
\newblock


\bibitem[\protect\citeauthoryear{Ely, Osheroff, Gorman, Ebell, Chambliss,
  Pifer, and Stavri}{Ely et~al\mbox{.}}{2000}]%
        {ely00}
{J.~W. Ely}, {J. Osheroff}, {P. Gorman}, {M. Ebell}, {L. Chambliss}, {E.
  Pifer}, {and} {Z. Stavri}. 2000.
\newblock \showarticletitle{A taxonomy of generic clinical questions:
  classification study}.
\newblock {\em BMJ\/}  {321} (2000), 429--432.
\newblock


\bibitem[\protect\citeauthoryear{Ely, Osheroff, Chambliss, Ebell, and
  Rosenbaum}{Ely et~al\mbox{.}}{2005}]%
        {ely05}
{J.~W. Ely}, {J.~A. Osheroff}, {M.~L. Chambliss}, {M.~H. Ebell}, {and} {M.~E.
  Rosenbaum}. 2005.
\newblock \showarticletitle{Answering Physicians' clinical Questions: Obstacles
  and Potential Solutions}.
\newblock {\em {Journal of the American Medical Informatics Association}\/}
  {12}, 2 (2005), 217--224.
\newblock


\bibitem[\protect\citeauthoryear{Ely, Osheroff, Ebell, Bergus, Levy, Chambliss,
  and Evans}{Ely et~al\mbox{.}}{1999}]%
        {ely99}
{J.~W. Ely}, {J.~A. Osheroff}, {M.~H. Ebell}, {G.~R. Bergus}, {B.~T. Levy},
  {M.~L. Chambliss}, {and} {E.~R. Evans}. 1999.
\newblock \showarticletitle{Analysis of questions asked by family doctors
  regarding patient care.}
\newblock {\em BMJ\/} {319}, 7206 (1999), 358--361.
\newblock


\bibitem[\protect\citeauthoryear{Ely, Osheroff, Ebell, Chambliss, Vinson,
  Stevermer, and Pifer}{Ely et~al\mbox{.}}{2002}]%
        {ely02}
{J.~W. Ely}, {J.~A. Osheroff}, {M.~H. Ebell}, {M.~L. Chambliss}, {D.~C.
  Vinson}, {J.~J. Stevermer}, {and} {E.~A. Pifer}. 2002.
\newblock \showarticletitle{Obstacles to Answering Doctors' Questions about
  Patient Care with Evidence: Qualitative Study}.
\newblock {\em BMJ\/} {324}, 7339 (2002), 710.
\newblock


\bibitem[\protect\citeauthoryear{Erkan and Radev}{Erkan and Radev}{2004}]%
        {erkan04}
{G. Erkan} {and} {D.~R. Radev}. 2004.
\newblock \showarticletitle{LexRank: Graph-based Centrality as Salience in Text
  Summarization}.
\newblock {\em Journal of Artificial Intelligence Research\/}  {22} (2004),
  457--479.
\newblock


\bibitem[\protect\citeauthoryear{Farzindar and Lapalme}{Farzindar and
  Lapalme}{2004}]%
        {farzinder04}
{A. Farzindar} {and} {G. Lapalme}. 2004.
\newblock \showarticletitle{Legal text summarization by exploration of the
  thematic structures and argumentative roles}. In {\em Text Summarization
  Branches Out Conference held in conjunction with ACL 2004}. 27--38.
\newblock


\bibitem[\protect\citeauthoryear{Fiszman, Demner-Fushman, Kilicoglu, and
  Rindflesch}{Fiszman et~al\mbox{.}}{2009}]%
        {fiszman09}
{M. Fiszman}, {D. Demner-Fushman}, {H. Kilicoglu}, {and} {T.~C. Rindflesch}.
  2009.
\newblock \showarticletitle{Automatic summarization of MEDLINE citations for
  evidence-based medical treatment: A topic-oriented evaluation}.
\newblock {\em Journal of Biomedical Informatics\/} {42}, 5 (2009), 801--813.
\newblock


\bibitem[\protect\citeauthoryear{Fiszman, Rindflesch, and Kilicoglu}{Fiszman
  et~al\mbox{.}}{2004}]%
        {fiszman04}
{M. Fiszman}, {T.~C. Rindflesch}, {and} {H. Kilicoglu}. 2004.
\newblock \showarticletitle{Abstraction Summarization for Managing the
  Biomedical Research Literature}. In {\em In Proceedings of the NAACL-HLT
  workshop on Computational Lexical Semantics}. 76--83.
\newblock


\bibitem[\protect\citeauthoryear{Fiszman, Rindflesch, and Kilicoglu}{Fiszman
  et~al\mbox{.}}{2003}]%
        {fiszman03}
{M.~A. Fiszman}, {T.~C. Rindflesch}, {and} {H. Kilicoglu}. 2003.
\newblock \showarticletitle{Integrating a hypernymic proposition interpreter
  into a semantic processor for biomedical texts.}
\newblock {\em {Proceedings of the AMIA Annual Symposium}\/} (2003), 239--243.
\newblock


\bibitem[\protect\citeauthoryear{Friedman}{Friedman}{2005}]%
        {friedman05}
{C. Friedman}. 2005.
\newblock {\em Knowledge management and datamining in biomedicine}.
\newblock Springer New York, Chapter Semantic Text Parsing for patient records,
  423--448.
\newblock


\bibitem[\protect\citeauthoryear{Friedman and Hripcsak}{Friedman and
  Hripcsak}{1999}]%
        {friedman99}
{C. Friedman} {and} {G. Hripcsak}. 1999.
\newblock \showarticletitle{Natural Language Processing and Its Future in
  Medicine}.
\newblock {\em Academic Medicine\/} {74}, 8 (August 1999), 890--893.
\newblock


\bibitem[\protect\citeauthoryear{Gaizauskas, Herring, Oakes, Beaulieu, Willett,
  Fowkes, and Jonsson}{Gaizauskas et~al\mbox{.}}{2001}]%
        {gaizauskas01}
{R. Gaizauskas}, {P. Herring}, {M. Oakes}, {M. Beaulieu}, {P. Willett}, {H.
  Fowkes}, {and} {A. Jonsson}. 2001.
\newblock \showarticletitle{Intelligent Access to Text: Integrating Information
  Extraction Technology into Text Browsers}. In {\em Proceedings of the Human
  Language Technology Conference (HLT)}.
\newblock


\bibitem[\protect\citeauthoryear{Gilbody}{Gilbody}{1996}]%
        {gilbody96}
{S. Gilbody}. 1996.
\newblock \showarticletitle{Evidence-based Medicine. An Improved Format for
  Journal Clubs}.
\newblock {\em Psychiatric Bulletin\/}  {20} (1996), 673--675.
\newblock


\bibitem[\protect\citeauthoryear{Godlee}{Godlee}{1998}]%
        {godlee99}
{F. Godlee}. 1998.
\newblock \showarticletitle{Getting evidence into practice}.
\newblock {\em BMJ\/}  {317} (1998), 6.
\newblock


\bibitem[\protect\citeauthoryear{Gorman and Helfand}{Gorman and
  Helfand}{1995}]%
        {gorman95}
{P.~N. Gorman} {and} {M. Helfand}. 1995.
\newblock \showarticletitle{{Information seeking in primary care: How
  physicians choose which clinical questions to pursue and which to leave
  unanswered}}.
\newblock {\em Medical Decision Making\/} {15}, 2 (1995), 113--119.
\newblock


\bibitem[\protect\citeauthoryear{Green and Ruff}{Green and Ruff}{2005}]%
        {green05}
{M.~L. Green} {and} {T.~R. Ruff}. 2005.
\newblock \showarticletitle{Why Do Residents Fail to Answer Their Clinical
  Questions? A Qualitative Study of Barriers to Practicing Evidence-Based
  Medicine}.
\newblock {\em Academic Medicine: Journal of the Association of American
  Medical Colleges\/} {80}, 2 (February 2005), 176--182.
\newblock


\bibitem[\protect\citeauthoryear{Greenhalgh}{Greenhalgh}{1999}]%
        {greenhalgh99}
{T. Greenhalgh}. 1999.
\newblock \showarticletitle{Narrative based medicine in an evidence based
  world}.
\newblock {\em BMJ\/}  {318} (1999), 323--325.
\newblock


\bibitem[\protect\citeauthoryear{Greenhalgh}{Greenhalgh}{2006}]%
        {greenhalgh06}
{T. Greenhalgh}. 2006.
\newblock {\em How to read a paper: The Basics of Evidence-based Medicine\/} (3
  ed.).
\newblock Blackwell Publishing.
\newblock


\bibitem[\protect\citeauthoryear{Hahn, Romacker, and Schulz}{Hahn
  et~al\mbox{.}}{2002}]%
        {hahn02}
{U. Hahn}, {M. Romacker}, {and} {S. Schulz}. 2002.
\newblock \showarticletitle{{MEDSYNDIKATE --- a natural language system for the
  extraction of medical information from findings reports}}.
\newblock {\em International Journal of Medical Informatics\/} {67}, 1--3
  (2002), 63--74.
\newblock


\bibitem[\protect\citeauthoryear{Hahn and Strube}{Hahn and Strube}{1997}]%
        {hahn97}
{U. Hahn} {and} {M. Strube}. 1997.
\newblock \showarticletitle{Centering in-the-large computing referential
  discourse segments}. In {\em Proceedings of the 35th Annual Meeting of the
  ACL and the 8th Conference of the European Chapter of the ACL}. 104--111.
\newblock


\bibitem[\protect\citeauthoryear{Harabagiu and Lacatusu}{Harabagiu and
  Lacatusu}{2005}]%
        {harabagiu05}
{S. Harabagiu} {and} {F. Lacatusu}. 2005.
\newblock \showarticletitle{Topic Themes for Multi-document Summarization}. In
  {\em Proceedings of the International ACM-SIGIR Conference on Research and
  Development in Information Retrieval}. 202--209.
\newblock


\bibitem[\protect\citeauthoryear{Haynes, Wilczynski, McKibbon, Walker, and
  Sinclair}{Haynes et~al\mbox{.}}{1994}]%
        {haynes94}
{R.~B. Haynes}, {N. Wilczynski}, {K.~A. McKibbon}, {C.~J. Walker}, {and} {J.~C.
  Sinclair}. 1994.
\newblock \showarticletitle{{Developing Optimal Search Strategies for Detecting
  Clinically Sound Studies in MEDLINE}}.
\newblock {\em {Journal of the American Medical Informatics Association}\/}
  {1}, 6 (1994), 447--458.
\newblock


\bibitem[\protect\citeauthoryear{Hearst}{Hearst}{1994}]%
        {hearst94}
{M.~A. Hearst}. 1994.
\newblock \showarticletitle{Multi-paragraph segmentation of expository text}.
  In {\em Proceedings of the 32nd Annual Meeting of the ACL}. Association for
  Computational Linguistics, 9--16.
\newblock


\bibitem[\protect\citeauthoryear{Heppin and Jarvelin}{Heppin and
  Jarvelin}{2012}]%
        {heppin12}
{K.~F. Heppin} {and} {A. Jarvelin}. 2012.
\newblock \showarticletitle{{Towards Improving Search Results for Medical
  Experts and Laypersons}}. In {\em {Proceedings of CLEFeHealth}}.
\newblock


\bibitem[\protect\citeauthoryear{Hersh, Crabtree, Hickman, Scherek, Friedman,
  Tidmarsh, Mosbaek, and Kraemer}{Hersh et~al\mbox{.}}{2002}]%
        {hersh02}
{W.~R. Hersh}, {M.~K. Crabtree}, {D.~H. Hickman}, {L. Scherek}, {C.~P.
  Friedman}, {P. Tidmarsh}, {C. Mosbaek}, {and} {D. Kraemer}. 2002.
\newblock \showarticletitle{{Factors associated with searching MEDLINE and
  applying evidence to answer clinical questions}}.
\newblock {\em {Journal of the American Medical Informatics Association}\/}
  {9} (2002), 283--293.
\newblock


\bibitem[\protect\citeauthoryear{Hoogendam, Stalenhoef, de~V.~Robb\'e, and
  Overbeke}{Hoogendam et~al\mbox{.}}{2008}]%
        {hoogendam08}
{A. Hoogendam}, {A.~F.~H. Stalenhoef}, {P.~F. de V.~Robb\'e}, {and} {A.~J.
  P.~M. Overbeke}. 2008.
\newblock \showarticletitle{Analysis of queries sent to PubMed at the point of
  care: observation of search behaviour in a medical teaching hospital.}
\newblock {\em BMC Medical Informatics and Decision Making\/}  {8} (2008), 42.
\newblock


\bibitem[\protect\citeauthoryear{Hovy and Lin}{Hovy and Lin}{1999}]%
        {hovy99}
{E. Hovy} {and} {C. Lin}. 1999.
\newblock {\em Advances in Automatic Text Summarization}.
\newblock MIT Press, Chapter Automated Text Summarisation in SUMMARIST, 81--94.
\newblock


\bibitem[\protect\citeauthoryear{Huang, Lin, and Demner-Fushman}{Huang
  et~al\mbox{.}}{2006}]%
        {huang06}
{X. Huang}, {J. Lin}, {and} {D. Demner-Fushman}. 2006.
\newblock \showarticletitle{Evaluation of {PICO} as a Knowledge Representation
  for Clinical Questions}. In {\em {Proceedings of AMIA Annual Symposium}}.
  359--363.
\newblock


\bibitem[\protect\citeauthoryear{Hunt and McKibbon}{Hunt and McKibbon}{1997}]%
        {hunt97}
{D.~L. Hunt} {and} {K.~A. McKibbon}. 1997.
\newblock \showarticletitle{Locating and Appraising Systematic Reviews}.
\newblock {\em Annals of Internal Medicine\/} {126}, 7 (1997), 532--538.
\newblock


\bibitem[\protect\citeauthoryear{Kabadjov, Steinberger, Steinberger, Poesio,
  and Pouliquen}{Kabadjov et~al\mbox{.}}{2010}]%
        {kabadjov10}
{M. Kabadjov}, {J. Steinberger}, {R. Steinberger}, {M. Poesio}, {and} {B.
  Pouliquen}. 2010.
\newblock \showarticletitle{Enhancing N-Gram-Based Summary Evaluation Using
  Information Content and a Taxonomy}.
\newblock In {\em Advances in Information Retrieval}. Lecture Notes in Computer
  Science, Vol. 5993. Springer Berlin / Heidelberg, 662--666.
\newblock


\bibitem[\protect\citeauthoryear{Karimi, Zobel, Pohl, and Scholer}{Karimi
  et~al\mbox{.}}{2009}]%
        {karimi09}
{S. Karimi}, {J. Zobel}, {S. Pohl}, {and} {F. Scholer}. 2009.
\newblock \showarticletitle{The Challenge of High Recall in Biomedical
  Systematic Search}. In {\em Proceedings of the third international workshop
  on Data and text mining in bioinformatics}. 89--92.
\newblock


\bibitem[\protect\citeauthoryear{Kelly, Dungs, Kriewel, Hanbury, Goeuriot,
  Jones, Langs, and Muller}{Kelly et~al\mbox{.}}{2014}]%
        {kelly14}
{L. Kelly}, {S. Dungs}, {S. Kriewel}, {A. Hanbury}, {L. Goeuriot}, {G.~J.~F.
  Jones}, {G. Langs}, {and} {H. Muller}. 2014.
\newblock \showarticletitle{{Professional: Multilingual, Multimodal
  Professional Medical Search}}. In {\em {ECIR 2014}}. 754 -- 758.
\newblock


\bibitem[\protect\citeauthoryear{Kim, Martinez, Cavedon, and Yencken}{Kim
  et~al\mbox{.}}{2011}]%
        {kim11}
{S.~N. Kim}, {D. Martinez}, {L. Cavedon}, {and} {L. Yencken}. 2011.
\newblock \showarticletitle{{Automatic classification of sentences to support
  Evidence Based Medicine.}}
\newblock {\em BMC bioinformatics\/}  {12 Suppl 2} (2011).
\newblock


\bibitem[\protect\citeauthoryear{Kupiec, Pedersen, and Chen}{Kupiec
  et~al\mbox{.}}{1995}]%
        {kupiec95}
{J. Kupiec}, {J. Pedersen}, {and} {F. Chen}. 1995.
\newblock \showarticletitle{A Trainable Document Summarizer}. In {\em
  Proceedings of the International ACM-SIGIR Conference on Research and
  Development in Information Retrieval}. 68--73.
\newblock


\bibitem[\protect\citeauthoryear{Lacatusu, Parker, and Harabagiu}{Lacatusu
  et~al\mbox{.}}{2003}]%
        {lacatusu03}
{F. Lacatusu}, {P. Parker}, {and} {S. Harabagiu}. 2003.
\newblock \showarticletitle{Lite-GISTexter: Generating Short Summaries with
  Minimal Resources}. In {\em Proceedings of the Document Understanding
  Conference}. 122--128.
\newblock


\bibitem[\protect\citeauthoryear{Lee, Cimino, Zhu, Sable, Shanker, Ely, and
  Yu}{Lee et~al\mbox{.}}{2006a}]%
        {lee06a}
{M. Lee}, {J. Cimino}, {H.~R. Zhu}, {C. Sable}, {V. Shanker}, {J. Ely}, {and}
  {H. Yu}. 2006a.
\newblock \showarticletitle{Beyond Information Retrieval -- Medical Question
  Answering}. In {\em {Proceedings of AMIA Annual Symposium}}. 469--473.
\newblock


\bibitem[\protect\citeauthoryear{Lee, Wang, and Yu}{Lee et~al\mbox{.}}{2006b}]%
        {lee06b}
{M. Lee}, {W. Wang}, {and} {H. Yu}. 2006b.
\newblock \showarticletitle{Exploring supervised and unsupervised methods to
  detect topics in biomedical text}.
\newblock {\em BMC Bioinformatics\/}  {7} (2006), 140.
\newblock


\bibitem[\protect\citeauthoryear{Lenci, Bartolini, Calzolari, Agua, Busemann,
  Cartier, Chevreau, and Coch}{Lenci et~al\mbox{.}}{2002}]%
        {lenci02}
{A. Lenci}, {R. Bartolini}, {N. Calzolari}, {A. Agua}, {S. Busemann}, {E.
  Cartier}, {K. Chevreau}, {and} {J. Coch}. 2002.
\newblock \showarticletitle{{Multilingual Summarization by Integrating
  Linguistic Resources in the MLIS-MUSI Project}}. In {\em Proceedings of the
  Third International Conference on Language Resources and Evaluation}.
  1464--1471.
\newblock


\bibitem[\protect\citeauthoryear{Leskovec, Milic-Frayling, and
  Grobelnik}{Leskovec et~al\mbox{.}}{2005}]%
        {leskovec05}
{J. Leskovec}, {N. Milic-Frayling}, {and} {M. Grobelnik}. 2005.
\newblock \showarticletitle{Impact of Linguistic Analysis on the Semantic Graph
  Coverage and Learning of Document Extracts}. In {\em {Proceedings of AAAI}}.
  1069--1074.
\newblock


\bibitem[\protect\citeauthoryear{Lin}{Lin}{1999}]%
        {lin99}
{C. Lin}. 1999.
\newblock \showarticletitle{Training a Selection Function for Extraction}. In
  {\em Proceedings of the Eighteenth Annual International ACM Conference on
  Information and Knowledge Management (CIKM)}. 1--8.
\newblock


\bibitem[\protect\citeauthoryear{Lin}{Lin}{2004}]%
        {lin04}
{C. Lin}. 2004.
\newblock \showarticletitle{{ROUGE: A Package for Automatic Evaluation of
  Summaries}}. In {\em Proceedings of NAACL-HLT}.
\newblock


\bibitem[\protect\citeauthoryear{Lin, Cao, Gao, and Nie}{Lin
  et~al\mbox{.}}{2006}]%
        {lin06a}
{C. Lin}, {G. Cao}, {J. Gao}, {and} {J. Nie}. 2006.
\newblock \showarticletitle{{An information-theoretic approach to automatic
  evaluation of summaries}}. In {\em Proceedings of NAACL-HLT}. 463--470.
\newblock


\bibitem[\protect\citeauthoryear{Lin and Hovy}{Lin and Hovy}{1997}]%
        {lin97}
{C. Lin} {and} {E. Hovy}. 1997.
\newblock \showarticletitle{Identifying topics by position}. In {\em
  Proceedings of the Fifth conference on Applied Natural Language Processing}.
  283--290.
\newblock


\bibitem[\protect\citeauthoryear{Lin and Hovy}{Lin and Hovy}{2000}]%
        {lin00}
{C. Lin} {and} {E. Hovy}. 2000.
\newblock \showarticletitle{The Automated Acquisition of Topic Signatures for
  Text Summarization}. In {\em Proceedings of the 18th International Conference
  on Computational Linguistics}. 495--501.
\newblock


\bibitem[\protect\citeauthoryear{Lin and Hovy}{Lin and Hovy}{2002}]%
        {lin02}
{C. Lin} {and} {E. Hovy}. 2002.
\newblock \showarticletitle{Manual and automatic evaluation of summaries}. In
  {\em Proceedings of the ACL-02 Workshop on Automatic Summarization}. 45--51.
\newblock


\bibitem[\protect\citeauthoryear{Lin and Hovy}{Lin and Hovy}{2003}]%
        {lin03}
{C. Lin} {and} {E. Hovy}. 2003.
\newblock \showarticletitle{{Automatic Evaluation of Summaries Using N-gram
  Co-occurrence Statistics}}. In {\em Proceedings of NAACL-HLT}. 71--78.
\newblock


\bibitem[\protect\citeauthoryear{Lin and Demner-Fushman}{Lin and
  Demner-Fushman}{2006}]%
        {lin06}
{J.~J. Lin} {and} {D. Demner-Fushman}. 2006.
\newblock \showarticletitle{The role of knowledge in conceptual retrieval: a
  study in the domain of clinical medicine}. In {\em Proceedings of the
  International ACM-SIGIR Conference on Research and Development in Information
  Retrieval}. 99--106.
\newblock


\bibitem[\protect\citeauthoryear{Lindberg, Humphreys, and McCray}{Lindberg
  et~al\mbox{.}}{1993}]%
        {lindberg93}
{D.~A. Lindberg}, {B.~L. Humphreys}, {and} {A. McCray}. 1993.
\newblock \showarticletitle{The Unified Medical Language System}.
\newblock {\em Methods of Information in Medicine\/}  {32} (1993), 281--291.
\newblock


\bibitem[\protect\citeauthoryear{Litvak and Last}{Litvak and Last}{2008}]%
        {litvak08}
{M. Litvak} {and} {M. Last}. 2008.
\newblock \showarticletitle{{Graph-Based Keyword Extraction for Single-Document
  Summarization}}. In {\em Proceedings of the Workshop on Multi-source
  Multilingual Information Extraction and Summarization}. 17--24.
\newblock


\bibitem[\protect\citeauthoryear{Louis and Nenkova}{Louis and Nenkova}{2008}]%
        {louis08}
{A. Louis} {and} {A. Nenkova}. 2008.
\newblock {\em Automatic Summary Evaluation without Human Models}.
\newblock {T}echnical {R}eport.
\newblock


\bibitem[\protect\citeauthoryear{Luhn}{Luhn}{1958}]%
        {luhn58}
{H.P. Luhn}. 1958.
\newblock \showarticletitle{The Automatic Creation of Literature Abstracts}.
\newblock {\em IBM Journal\/}  {2} (1958), 159--165.
\newblock


\bibitem[\protect\citeauthoryear{Mani}{Mani}{2001}]%
        {mani01}
{I. Mani}. 2001.
\newblock {\em Automatic Summarization}.
\newblock John Benjamins Publishing Company.
\newblock


\bibitem[\protect\citeauthoryear{Mani and Bloedorn}{Mani and Bloedorn}{1997}]%
        {mani97}
{I. Mani} {and} {E. Bloedorn}. 1997.
\newblock \showarticletitle{Multi-Document Summarization by Graph Search and
  Matching}. In {\em {Proceedings of AAAI}}. 622--628.
\newblock


\bibitem[\protect\citeauthoryear{Mani, (editors, and Sanderson)}{Mani
  et~al\mbox{.}}{2000}]%
        {mani00}
{I. Mani}, {M.~T.~Maybury (editors}, {and} {M. Sanderson)}. 2000.
\newblock {\em Book Review: Advances in Automatic Text Summarization}.
\newblock MIT Press.
\newblock


\bibitem[\protect\citeauthoryear{Mani, Klein, House, Hirschman, Firmin, and
  Sundhem}{Mani et~al\mbox{.}}{2002}]%
        {mani02}
{I. Mani}, {G. Klein}, {D. House}, {L. Hirschman}, {T. Firmin}, {and} {B.
  Sundhem}. 2002.
\newblock \showarticletitle{SUMMAC: A text summarisation evaluation}.
\newblock {\em Natural Language Engineering\/} {8}, 1 (2002), 43--68.
\newblock


\bibitem[\protect\citeauthoryear{Mani and Maybury}{Mani and Maybury}{1999}]%
        {mani99}
{I. Mani} {and} {M.~T. Maybury}. 1999.
\newblock {\em Advances in Automatic Text Summarization}.
\newblock MIT Press.
\newblock


\bibitem[\protect\citeauthoryear{Marcu}{Marcu}{1998}]%
        {marcu98}
{D. Marcu}. 1998.
\newblock \showarticletitle{To Build Text Summaries of High Quality, Nuclearity
  is not Sufficient}. In {\em {Proceedings of AAAI}}. 1--8.
\newblock


\bibitem[\protect\citeauthoryear{Marcu}{Marcu}{1999}]%
        {marcu99}
{D. Marcu}. 1999.
\newblock {\em Advances in Automatic Text Summarization}.
\newblock MIT Press, Chapter Discourse Trees are Good Indicators of Importance
  in Text, 123--136.
\newblock


\bibitem[\protect\citeauthoryear{Marcu}{Marcu}{2000}]%
        {marcu00}
{D. Marcu}. 2000.
\newblock {\em The Theory and Practice of Discourse Parsing and Summarisation}.
\newblock Cambridge MA: MIT Press.
\newblock


\bibitem[\protect\citeauthoryear{McKeown, Barzilay, Evans, Hatzivassiloglou,
  Klavans, Nenkova, Sable, Schiffman, and Sigelman}{McKeown
  et~al\mbox{.}}{2002}]%
        {mckeown02}
{K.~R. McKeown}, {R. Barzilay}, {D. Evans}, {V. Hatzivassiloglou}, {J.~L.
  Klavans}, {A. Nenkova}, {C. Sable}, {B. Schiffman}, {and} {S. Sigelman}.
  2002.
\newblock \showarticletitle{Tracking and summarizing news on a daily basis with
  Columbia's Newsblaster}. In {\em Proceedings of the second international
  conference on Human Language Technology Research}. 280--285.
\newblock


\bibitem[\protect\citeauthoryear{McKeown, Jordan, and Hatzivassiloglou}{McKeown
  et~al\mbox{.}}{1998}]%
        {mckeown98}
{K.~R. McKeown}, {D.~A. Jordan}, {and} {V. Hatzivassiloglou}. 1998.
\newblock {\em Generating Patient-Specific Summaries of Online Literature}.
\newblock {T}echnical {R}eport. AAAI Technical Report SS-98-06. 34--43 pages.
\newblock


\bibitem[\protect\citeauthoryear{McKeown and Radev}{McKeown and Radev}{1995}]%
        {mckeown95a}
{K.~R. McKeown} {and} {D.~R. Radev}. 1995.
\newblock \showarticletitle{Generating Summaries of Multiple News Articles}. In
  {\em Proceedings of the International ACM-SIGIR Conference on Research and
  Development in Information Retrieval}. 74--82.
\newblock


\bibitem[\protect\citeauthoryear{Mihalcea}{Mihalcea}{2004}]%
        {mihalcea04}
{R. Mihalcea}. 2004.
\newblock \showarticletitle{{Graph-based ranking algorithms for sentence
  extraction, applied to summarization}}. In {\em Proceedings of the ACL 2004
  on Interactive poster and demonstration sessions}.
\newblock


\bibitem[\protect\citeauthoryear{Mihalcea and Tarau}{Mihalcea and
  Tarau}{2004}]%
        {mihalcea11}
{R. Mihalcea} {and} {P. Tarau}. 2004.
\newblock \showarticletitle{{TextRank: Bringing Order into Texts}}. In {\em
  Proceedings of EMNLP}. Barcelona, Spain.
\newblock


\bibitem[\protect\citeauthoryear{Miike, Itoh, Ono, and Sumita}{Miike
  et~al\mbox{.}}{1994}]%
        {miike94}
{S. Miike}, {E. Itoh}, {K. Ono}, {and} {K. Sumita}. 1994.
\newblock \showarticletitle{A Full-text Retrieval System with a Dynamic
  Abstract Generation Function}. In {\em Proceedings of the International
  ACM-SIGIR Conference on Research and Development in Information Retrieval}.
  152--161.
\newblock


\bibitem[\protect\citeauthoryear{Miller, Pople, and Myers}{Miller
  et~al\mbox{.}}{1982}]%
        {miller82}
{R.~A. Miller}, {H.~E. Pople}, {and} {J.~D. Myers}. 1982.
\newblock \showarticletitle{Internist-1, an experimental computer-based
  diagnostic consultant for general internal medicine}.
\newblock {\em The New England Journal of Medicine\/} {307}, 8 (1982),
  468--478.
\newblock


\bibitem[\protect\citeauthoryear{Moll{\'a}, Santiago-Mart{\'i}nez, Sarker, and
  Paris}{Moll{\'a} et~al\mbox{.}}{2015}]%
        {molla15}
{D. Moll{\'a}}, {M.~E. Santiago-Mart{\'i}nez}, {A. Sarker}, {and} {C. Paris}.
  2015.
\newblock \showarticletitle{A corpus for research in text processing for
  evidence based medicine}.
\newblock {\em Language Resources and Evaluation\/} (2015), 1--23.
\newblock
\showISSN{1574-0218}
\showDOI{%
\url{http://dx.doi.org/10.1007/s10579-015-9327-2}}


\bibitem[\protect\citeauthoryear{Moll{\'a} and Vicedo}{Moll{\'a} and
  Vicedo}{2007}]%
        {molla07}
{D. Moll{\'a}} {and} {J.~L. Vicedo}. 2007.
\newblock \showarticletitle{{Question Answering in Restricted Domains: An
  Overview}}.
\newblock {\em Computational Linguistics\/}  {33} (2007), 41--61.
\newblock


\bibitem[\protect\citeauthoryear{Montori, Wilczynski, Morgan, and
  Haynes}{Montori et~al\mbox{.}}{2005}]%
        {montori04}
{V.~M. Montori}, {N.~L. Wilczynski}, {D. Morgan}, {and} {R.~B. Haynes}. 2005.
\newblock \showarticletitle{Optimal search strategies for retrieving systematic
  reviews from Medline: analytical survey.}
\newblock {\em BMJ\/} {330}, 7482 (2005), 68--73.
\newblock


\bibitem[\protect\citeauthoryear{Morris, Kasper, and Adams}{Morris
  et~al\mbox{.}}{1992}]%
        {morris92}
{A.~H. Morris}, {G.~M. Kasper}, {and} {D.~A. Adams}. 1992.
\newblock \showarticletitle{{The Effects and Limitations of Automated Text
  Condensing on Reading Comprehension Performance}}.
\newblock {\em {Information Systems Research}\/} {3}, 1 (1992), 17--35.
\newblock


\bibitem[\protect\citeauthoryear{Nenkova and Passonneau}{Nenkova and
  Passonneau}{2004}]%
        {nenkova04}
{A. Nenkova} {and} {R. Passonneau}. 2004.
\newblock \showarticletitle{Evaluating Content Selection in Summarization: The
  Pyramid Method}. In {\em Proceedings of NAACL-HLT}.
\newblock


\bibitem[\protect\citeauthoryear{Niu and Hirst}{Niu and Hirst}{2004}]%
        {niu04}
{Y. Niu} {and} {G. Hirst}. 2004.
\newblock \showarticletitle{Analysis of semantic classes in medical text for
  question answering}. In {\em {Proceedings of the ACL-2004 workshop Question
  Answering in Restricted Domains, Barcelona, Spain}}.
\newblock


\bibitem[\protect\citeauthoryear{Niu, Hirst, McArthur, and
  Rodriguez-Gianolli}{Niu et~al\mbox{.}}{2003}]%
        {niu03}
{Y. Niu}, {G. Hirst}, {G. McArthur}, {and} {P. Rodriguez-Gianolli}. 2003.
\newblock \showarticletitle{{Answering Clinical Questions with Role
  Identification}}. In {\em {Proceedings of the ACL-2003 workshop Natural
  Language Processing in Biomedicine}}.
\newblock


\bibitem[\protect\citeauthoryear{Niu, Zhu, and Hirst}{Niu
  et~al\mbox{.}}{2006}]%
        {niu06}
{Y. Niu}, {X. Zhu}, {and} {G. Hirst}. 2006.
\newblock \showarticletitle{{Using outcome polarity in sentence extraction for
  medical question-answering}}. In {\em {Proceedings of AMIA Annual
  Symposium}}. 599--603.
\newblock


\bibitem[\protect\citeauthoryear{Niu, Zhu, Li, and Hirst}{Niu
  et~al\mbox{.}}{2005}]%
        {niu05}
{Y. Niu}, {X. Zhu}, {J. Li}, {and} {G. Hirst}. 2005.
\newblock \showarticletitle{Analysis of polarity information in medical text}.
  In {\em {Proceedings of AMIA Annual Symposium}}. 570--574.
\newblock


\bibitem[\protect\citeauthoryear{Osborne}{Osborne}{2002}]%
        {osborne02}
{M. Osborne}. 2002.
\newblock \showarticletitle{Using Maximum Entropy for Sentence Extraction}. In
  {\em Proceedings of the ACL'02 Workshop on Automatic Summarization}. 1--8.
\newblock


\bibitem[\protect\citeauthoryear{Papineni, Roukos, Ward, and Zhu}{Papineni
  et~al\mbox{.}}{2002}]%
        {papineni02}
{K. Papineni}, {S. Roukos}, {T. Ward}, {and} {W.~J. Zhu}. 2002.
\newblock \showarticletitle{BLEU: A Method for Aytomatic Evaluation of Machine
  Translation}. In {\em Proceedings of the 40th Annual Meeting of the ACL}.
\newblock


\bibitem[\protect\citeauthoryear{Plaza, Diaz, and Gervas}{Plaza
  et~al\mbox{.}}{2011}]%
        {plaza11a}
{L. Plaza}, {A. Diaz}, {and} {P. Gervas}. 2011.
\newblock \showarticletitle{A semantic graph-based approach to biomedical
  summarisation}.
\newblock {\em Artificial Intelligence in Medicine\/}  {53} (2011), 1--14.
\newblock


\bibitem[\protect\citeauthoryear{Pohl, Zobel, and Moffat}{Pohl
  et~al\mbox{.}}{2010}]%
        {pohl10}
{S. Pohl}, {J. Zobel}, {and} {A. Moffat}. 2010.
\newblock \showarticletitle{Extended Boolean Retrieval for Systematic
  Biomedical Reviews}. In {\em Proceedings of the thirty-third Australasian
  Computer Science Conference}.
\newblock


\bibitem[\protect\citeauthoryear{Polanyi, Culy, van~den Berg, Thione, and
  Ahn}{Polanyi et~al\mbox{.}}{2004}]%
        {polanyi04}
{L. Polanyi}, {C. Culy}, {M. van~den Berg}, {G.~L. Thione}, {and} {D. Ahn}.
  2004.
\newblock \showarticletitle{A Rule-based Approach to Discourse Parsing}. In
  {\em Proceedings of the fifth SIGdial workshop on Discourse and Dialogue}.
  108--117.
\newblock


\bibitem[\protect\citeauthoryear{Radev, Hovy, and McKeown}{Radev
  et~al\mbox{.}}{2002}]%
        {radev02}
{D.~R. Radev}, {E. Hovy}, {and} {K. McKeown}. 2002.
\newblock \showarticletitle{Introduction to the special issue on
  summarization}.
\newblock {\em Computational Linguistics\/} {28}, 4 (2002), 399--408.
\newblock


\bibitem[\protect\citeauthoryear{Radev, Jing, and Budzikowska}{Radev
  et~al\mbox{.}}{2000}]%
        {radev00a}
{D.~R. Radev}, {H. Jing}, {and} {M. Budzikowska}. 2000.
\newblock \showarticletitle{Centroid-based Summarization of Multiple Documents:
  Sentence Extraction, Utility-based Evaluation, and User Studies}. In {\em
  Proceedings of the NAACL-ANLP Workshop on Automatic Summarization}. 21--30.
\newblock


\bibitem[\protect\citeauthoryear{Radev, Jing, Stys, and Tam}{Radev
  et~al\mbox{.}}{2004}]%
        {radev04}
{D.~R. Radev}, {H. Jing}, {M. Stys}, {and} {D. Tam}. 2004.
\newblock \showarticletitle{Centroid-based Summarization of Multiple
  Documents}.
\newblock {\em Information Processing and Management\/}  {40} (2004), 919--938.
\newblock


\bibitem[\protect\citeauthoryear{Radev and Mckeown}{Radev and Mckeown}{1998}]%
        {radev98}
{D.~R. Radev} {and} {K.~R. Mckeown}. 1998.
\newblock \showarticletitle{Generating Natural Language Summaries from Multiple
  On-line Sources}.
\newblock {\em Computational Linguistics\/} {24}, 3 (1998), 470--500.
\newblock


\bibitem[\protect\citeauthoryear{Radev, Prager, and Samn}{Radev
  et~al\mbox{.}}{2000}]%
        {radev00}
{D.~R. Radev}, {J.~M. Prager}, {and} {V. Samn}. 2000.
\newblock \showarticletitle{Ranking suspected answers to natural language
  questions using predictive annotation}. In {\em Proceedings of the sixth
  conference on Applied Natural Language Processing}. 150--157.
\newblock


\bibitem[\protect\citeauthoryear{Reeve, Han, and Brooks}{Reeve
  et~al\mbox{.}}{2006}]%
        {reeve06a}
{L.~H. Reeve}, {H. Han}, {and} {A.~D. Brooks}. 2006.
\newblock \showarticletitle{BioChain: Using lexical chaining methods for
  biomedical text summarization}. In {\em Proceedings of the 21st annual ACM
  symposium on applied computing, bioinformatics track}. 180--184.
\newblock


\bibitem[\protect\citeauthoryear{Reeve, Han, and Brooks}{Reeve
  et~al\mbox{.}}{2007}]%
        {reeve07}
{L.~H. Reeve}, {H. Han}, {and} {A.~D. Brooks}. 2007.
\newblock \showarticletitle{The use of domain-specific concepts in biomedical
  text summarization}.
\newblock {\em Information Processing and Management\/}  {43} (2007),
  1765--1776.
\newblock


\bibitem[\protect\citeauthoryear{Reeve, Han, Nagori, Yang, Schwimmer, and
  Brooks}{Reeve et~al\mbox{.}}{2006}]%
        {reeve06b}
{L.~H. Reeve}, {H. Han}, {S.~V. Nagori}, {J.~C. Yang}, {T.~A. Schwimmer}, {and}
  {A.~D. Brooks}. 2006.
\newblock \showarticletitle{Concept frequency distribution in biomedical text
  summarization}. In {\em Proceedings of the ACM 15th conference on information
  and knowledge management (CIKM'06)}. 604--611.
\newblock


\bibitem[\protect\citeauthoryear{Richardson, Wilson, Nishikawa, and
  Hayward}{Richardson et~al\mbox{.}}{1995}]%
        {richardson95}
{S.~W. Richardson}, {M.~C. Wilson}, {J. Nishikawa}, {and} {R.~S. Hayward}.
  1995.
\newblock \showarticletitle{{The well-built clinical question: a key to
  evidence-based decisions}}.
\newblock {\em ACP Journal Club\/} {123}, 3 (1995), A12--A13.
\newblock


\bibitem[\protect\citeauthoryear{Rindflesch, Fiszman, and Libbus}{Rindflesch
  et~al\mbox{.}}{2005}]%
        {rindflesch05}
{T.~C. Rindflesch}, {M. Fiszman}, {and} {B. Libbus}. 2005.
\newblock Chapter 14 SEMANTIC INTERPRETATION FOR THE BIOMEDICAL RESEARCH
  LITERATURE.
\newblock   (2005).
\newblock


\bibitem[\protect\citeauthoryear{Rosenberg and Donald}{Rosenberg and
  Donald}{1995}]%
        {rosenberg95}
{W. Rosenberg} {and} {A. Donald}. 1995.
\newblock \showarticletitle{{Evidence based medicine: an approach to clinical
  problem-solving}}.
\newblock {\em BMJ\/}  {310} (1995), 1122--1126.
\newblock


\bibitem[\protect\citeauthoryear{Sackett, Rosenberg, Gray, Haynes, and
  Richardson}{Sackett et~al\mbox{.}}{1996}]%
        {sacket96}
{D.~L. Sackett}, {W.~M.~C. Rosenberg}, {J.~A.~M. Gray}, {B.~R. Haynes}, {and}
  {W.~S. Richardson}. 1996.
\newblock \showarticletitle{Evidence based medicine: what it is and what it
  isn't}.
\newblock {\em BMJ\/} {312}, 7023 (1996), 71--72.
\newblock
\showURL{%
\url{http://www.bmj.com}}


\bibitem[\protect\citeauthoryear{Sager, Laman, Bucknall, Nhan, and Tick}{Sager
  et~al\mbox{.}}{1994}]%
        {sager94}
{N. Sager}, {M. Laman}, {C. Bucknall}, {N. Nhan}, {and} {L.~J. Tick}. 1994.
\newblock \showarticletitle{Natural Language Processing and the Representation
  of Clinical Data}.
\newblock {\em {Journal of the American Medical Informatics Association}\/}
  {1} (1994), 142--160.
\newblock


\bibitem[\protect\citeauthoryear{Saggion and Lapalme}{Saggion and
  Lapalme}{2002}]%
        {saggion02}
{H. Saggion} {and} {G. Lapalme}. 2002.
\newblock \showarticletitle{Generating indicative-informative summaries with
  sumUM}.
\newblock {\em Computational Linguistics\/} {28}, 4 (2002), 497--526.
\newblock


\bibitem[\protect\citeauthoryear{Sarker}{Sarker}{2014}]%
        {sarker14thesis}
{A. Sarker}. 2014.
\newblock {\em {Automated Medical Text Summarisation to Support Evidence-based
  Medicine}}.
\newblock Ph.D. Dissertation. Macquarie University.
\newblock
\showURL{%
\url{http://web.science.mq.edu.au/~diego/theses/AbeedSarker.pdf}}


\bibitem[\protect\citeauthoryear{Sarker, Moll\'a, and Paris}{Sarker
  et~al\mbox{.}}{2012}]%
        {sarker12alta}
{A. Sarker}, {D. Moll\'a}, {and} {C. Paris}. 2012.
\newblock \showarticletitle{{Towards Two-step Multi-Document Summarisation for
  Evidence Based Medicine}}. In {\em Proceedings of the ALTW}. 79--87.
\newblock


\bibitem[\protect\citeauthoryear{Sarker, Moll\'a, and Paris}{Sarker
  et~al\mbox{.}}{2013a}]%
        {sarker13aime}
{A. Sarker}, {D. Moll\'a}, {and} {C. Paris}. 2013a.
\newblock \showarticletitle{An Approach for Query-Focused Text Summarisation
  for Evidence Based Medicine}.
\newblock In {\em Artificial Intelligence in Medicine}, {N.~Peek},
  {R.~M.~Morales}, {and} {M.~Peleg} (Eds.). Lecture Notes in Computer Science,
  Vol. 7885. Springer Berlin Heidelberg, 295--304.
\newblock
\showISBNx{978-3-642-38325-0}


\bibitem[\protect\citeauthoryear{Sarker, Moll\'a, and Paris}{Sarker
  et~al\mbox{.}}{2013b}]%
        {sarker13ijcnlp}
{A. Sarker}, {D. Moll\'a}, {and} {C. Paris}. 2013b.
\newblock \showarticletitle{{Automatic Prediction of Evidence-based
  Recommendations via Sentence-level Polarity Classification}}. In {\em
  {Proceedings of the International Joint Conference on Natural Language
  Processing}}. 712--718.
\newblock


\bibitem[\protect\citeauthoryear{Sarker, Moll\'a, and Paris}{Sarker
  et~al\mbox{.}}{2015}]%
        {sarker15aiim}
{A. Sarker}, {D. Moll\'a}, {and} {C. Paris}. 2015.
\newblock \showarticletitle{Automatic evidence quality prediction to support
  evidence-based decision making}.
\newblock {\em {Artificial Intelligence in Medicine}\/} {64}, 2 (June 2015),
  89--103.
\newblock


\bibitem[\protect\citeauthoryear{Sarker, Moll\'a, and Paris}{Sarker
  et~al\mbox{.}}{2016}]%
        {sarker16summ}
{A. Sarker}, {D. Moll\'a}, {and} {C. Paris}. 2016.
\newblock \showarticletitle{Query-oriented evidence extraction to support
  evidence-based medicine practice}.
\newblock {\em {Journal of Biomedical Informatics}\/}  {59} (2016), 169--184.
\newblock


\bibitem[\protect\citeauthoryear{Sauper and Barzilay}{Sauper and
  Barzilay}{2009}]%
        {sauper09}
{C. Sauper} {and} {R. Barzilay}. 2009.
\newblock \showarticletitle{Automatically Generating Wikipedia Articles: A
  Structure-Aware Approach}. In {\em Proceedings of the ACL}.
\newblock


\bibitem[\protect\citeauthoryear{Schilder and Kondadadi}{Schilder and
  Kondadadi}{2008}]%
        {schilder08}
{F. Schilder} {and} {R. Kondadadi}. 2008.
\newblock \showarticletitle{FastSum: Fast and Accurate Query-based
  Multi-document Summarization}. In {\em Proceedings of ACL-HLT, Short Papers}.
  205--208.
\newblock


\bibitem[\protect\citeauthoryear{Scholosser, Koul, and Costello}{Scholosser
  et~al\mbox{.}}{2006}]%
        {schlosser06}
{R.~W. Scholosser}, {R. Koul}, {and} {J. Costello}. 2006.
\newblock \showarticletitle{Asking well-built questions for evidence-based
  practice in augmentative and alternative communication}.
\newblock {\em Journal of Communication Disorders\/}  {40} (2006), 225--238.
\newblock


\bibitem[\protect\citeauthoryear{Schwitter}{Schwitter}{2010}]%
        {schwitter10}
{R. Schwitter}. 2010.
\newblock \showarticletitle{{Creating and Querying Formal Ontologies via
  Controlled Natural Language}}.
\newblock {\em Applied Artificial Intelligence\/} {24}, 1-2 (2010), 149--174.
\newblock


\bibitem[\protect\citeauthoryear{Selvaraj, Kumar, E., Saraswathi, Balaji,
  Nagamani, and Mohan}{Selvaraj et~al\mbox{.}}{2010}]%
        {selvaraj10}
{S. Selvaraj}, {Y. Kumar}, {E.}, {P. Saraswathi}, {D. Balaji}, {P. Nagamani},
  {and} {S.~K. Mohan}. 2010.
\newblock \showarticletitle{Evidence-based medicine - a new approach to teach
  medicine: a basic review for beginners}.
\newblock {\em Biology and Medicine\/} {2}, 1 (2010), 1--5.
\newblock


\bibitem[\protect\citeauthoryear{Shonjania and Bero}{Shonjania and
  Bero}{2001}]%
        {shonjania01}
{K.~G. Shonjania} {and} {L.~A. Bero}. 2001.
\newblock \showarticletitle{Taking Advantage of the Explosion of Systematic
  Reviews: An Efficient MEDLINE Search Strategy}.
\newblock {\em Effective Clinical Practice\/} {4}, 4 (2001), 157--162.
\newblock


\bibitem[\protect\citeauthoryear{Shortliffe}{Shortliffe}{1990}]%
        {shortliffe90}
{E.~H. Shortliffe}. 1990.
\newblock \showarticletitle{{Computer Programs to Support Clinical Decision
  Making}}.
\newblock {\em The Journal of the American Medical Association\/} {258}, 1
  (1990), 61--66.
\newblock


\bibitem[\protect\citeauthoryear{Shortliffe, Buchanan, and
  Feigenbaum}{Shortliffe et~al\mbox{.}}{1979}]%
        {shortliffe79}
{E.~H. Shortliffe}, {B.~G. Buchanan}, {and} {E.~A. Feigenbaum}. 1979.
\newblock \showarticletitle{Knowledge Engineering for Medical Decision Making:
  A Review of Computer-based Clinical Decision Aids}. In {\em Proceedings of
  the IEEE}, Vol.~67. 1207--1223.
\newblock


\bibitem[\protect\citeauthoryear{Slawson and Shaughnessy}{Slawson and
  Shaughnessy}{2005}]%
        {slawson05}
{D.~C. Slawson} {and} {A.~F. Shaughnessy}. 2005.
\newblock \showarticletitle{Teaching Evidence-Based Medicine: Should We Be
  Teaching Information Management Instead?}
\newblock {\em Academic Medicine\/} {80}, 7 (2005), 685--689.
\newblock


\bibitem[\protect\citeauthoryear{{Sparck Jones}}{{Sparck Jones}}{1999}]%
        {sparck99}
{K. {Sparck Jones}}. 1999.
\newblock \showarticletitle{Automatic summarizing: factors and directions}.
\newblock In {\em Advances in automatic text summarization}, {Inderjeet Mani}
  {and} {Mark~T. Maybury} (Eds.). The MIT Press, Chapter~1, 1 -- 12.
\newblock


\bibitem[\protect\citeauthoryear{{Sparck Jones}}{{Sparck Jones}}{2007}]%
        {sparck07}
{K. {Sparck Jones}}. 2007.
\newblock \showarticletitle{Automatic summarising: The state of the art}.
\newblock {\em Information Processing and Management\/}  {43} (2007), 1449 --
  1481.
\newblock


\bibitem[\protect\citeauthoryear{Suominen, Pyysalo, Hissa, Ginter, Liu, and
  et~al.}{Suominen et~al\mbox{.}}{2008}]%
        {suominen08}
{H. Suominen}, {S. Pyysalo}, {M. Hissa}, {F. Ginter}, {S. Liu}, {and}
  {D.~Marghescu et al.} 2008.
\newblock {\em {Handbook of Research on Text and Web Mining Technologies}}.
\newblock {IGI Global}, Chapter {Performance Evaluation Measures for Text
  Mining}, 724 -- 747.
\newblock


\bibitem[\protect\citeauthoryear{Suominen, Salantera, Velupillai, Chapman,
  Savova, Elhadad, Pradhan, South, Mowery, Jones, Leveling, Kelly, Goeuriot,
  Martinez, and Zuccon}{Suominen et~al\mbox{.}}{2013}]%
        {suominen13}
{H. Suominen}, {S. Salantera}, {S. Velupillai}, {W.~W. Chapman}, {G. Savova},
  {N. Elhadad}, {S. Pradhan}, {B.~R. South}, {D.~L. Mowery}, {G.~J.~F. Jones},
  {J. Leveling}, {L. Kelly}, {L. Goeuriot}, {D. Martinez}, {and} {G. Zuccon}.
  2013.
\newblock \showarticletitle{{Overview of the ShARe/CLEF eHealth Evaluation Lab
  2013}}.
\newblock In {\em {Information Access Evaluation. Multilinguality,
  Multimodality, and Visualization}}, {Pamela Forner}, {Henning Muller},
  {Roberto Predes}, {Paolo Rosso}, {and} {Benno Stein} (Eds.). {Lecture Notes
  in Computer Science}, Vol. 8138. 212--231.
\newblock


\bibitem[\protect\citeauthoryear{Svore, Vanderwende, and Burges}{Svore
  et~al\mbox{.}}{2007}]%
        {svore07}
{K.~M. Svore}, {L. Vanderwende}, {and} {C.~J.~C. Burges}. 2007.
\newblock \showarticletitle{Enhancing Single-document summarization by
  combining RankNet and Third-party Sources}. In {\em Proceedings of the 2007
  Joint Conference on EMNLP-CoNLL}. 448--457.
\newblock


\bibitem[\protect\citeauthoryear{Taylor, McAvoy, and O'Dowd}{Taylor
  et~al\mbox{.}}{2003}]%
        {taylor03}
{R.~J. Taylor}, {B.~R. McAvoy}, {and} {T. O'Dowd}. 2003.
\newblock {\em General Practice Medicine: an illustrated colour text}.
\newblock Elsevier Health Sciences.
\newblock


\bibitem[\protect\citeauthoryear{Teufel}{Teufel}{2001}]%
        {teufel01}
{S. Teufel}. 2001.
\newblock \showarticletitle{Task-Based Evaluation of Summary Quality:
  Describing Relationships between Scientific Papers}. In {\em Proceedings of
  the NAACL 2001 Workshop on Automatic Summarization}. 12--21.
\newblock


\bibitem[\protect\citeauthoryear{Teufel and Moens}{Teufel and Moens}{1997}]%
        {teufel97}
{S. Teufel} {and} {M. Moens}. 1997.
\newblock \showarticletitle{Sentence Extraction as a Classification Task}. In
  {\em Proceedings of the ACL-97}. 58--65.
\newblock


\bibitem[\protect\citeauthoryear{Thione, Van~den Berg, Polanyi, and
  Culy}{Thione et~al\mbox{.}}{2004}]%
        {thione04}
{G.~L. Thione}, {M. Van~den Berg}, {L. Polanyi}, {and} {C. Culy}. 2004.
\newblock \showarticletitle{Hybrid Text Summarization: Combining External
  Relevance Measures with Structural Analysis}. In {\em Proceedings of the Text
  Summarization Branches Out: Proceedings of the ACL-04 Workshop}. 51--55.
\newblock


\bibitem[\protect\citeauthoryear{Tutos and Moll{\'a}}{Tutos and
  Moll{\'a}}{2010}]%
        {tutos10}
{A. Tutos} {and} {D. Moll{\'a}}. 2010.
\newblock \showarticletitle{A Study on the Use of Search Engines for Answering
  Clinical Questions}. In {\em Proceedings of the Australian HIKM Workshop}.
  61--68.
\newblock


\bibitem[\protect\citeauthoryear{van Halteren and Teufel}{van Halteren and
  Teufel}{2003}]%
        {halteren03}
{H. van Halteren} {and} {S. Teufel}. 2003.
\newblock \showarticletitle{{Examining the concensus between human summaries:
  initial experiments}}. In {\em Proceedings of the NAACL-HLT Workshop on Text
  summarization}, Vol.~5. 57--64.
\newblock


\bibitem[\protect\citeauthoryear{Verhoeven, Boerma, and de~Jong}{Verhoeven
  et~al\mbox{.}}{2000}]%
        {verhoeven00}
{A.~A.~H. Verhoeven}, {E.~J. Boerma}, {and} {B.~M. de Jong}. 2000.
\newblock \showarticletitle{Which literature retrieval method is most effective
  for GPs?}
\newblock {\em Family Practice\/} {17}, 1 (2000), 30--35.
\newblock


\bibitem[\protect\citeauthoryear{Wang, Hu, Feng, and Wenyin}{Wang
  et~al\mbox{.}}{2007}]%
        {weiming07}
{W. Wang}, {D. Hu}, {M. Feng}, {and} {L. Wenyin}. 2007.
\newblock \showarticletitle{{Automatic clinical question answering based on
  UMLS relations}}. In {\em {Proceedings of the third International Conference
  on Semantics, Knowledge and Grid}}.
\newblock


\bibitem[\protect\citeauthoryear{Workman, Fiszman, and Hurdle}{Workman
  et~al\mbox{.}}{2012}]%
        {workman12}
{T.~E. Workman}, {M. Fiszman}, {and} {J.~F. Hurdle}. 2012.
\newblock \showarticletitle{Text summarisation as a decision support aid}.
\newblock {\em {BMC Medical Informatics and Decision Making}\/}  {12} (2012),
  41--53.
\newblock


\bibitem[\protect\citeauthoryear{Xu, Stenner, Doan, Johnson, Waltman, and
  Denny}{Xu et~al\mbox{.}}{2010}]%
        {xu10}
{H. Xu}, {S.~P. Stenner}, {S. Doan}, {K.~B. Johnson}, {L.~R. Waltman}, {and}
  {J.~C. Denny}. 2010.
\newblock \showarticletitle{{MedEx: a medication information extraction system
  for clinical narratives}}.
\newblock {\em {Journal of the American Medical Informatics Association}\/}
  {17} (2010), 19--24.
\newblock


\bibitem[\protect\citeauthoryear{Yih, Goodman, Vanderwende, and Suzuki}{Yih
  et~al\mbox{.}}{2007}]%
        {yih07}
{W. Yih}, {J. Goodman}, {L. Vanderwende}, {and} {H. Suzuki}. 2007.
\newblock \showarticletitle{Multi-Document Summarization by Maximizing
  Informative Content-Words}. In {\em Proceedings of the 20th International
  Joint Conference on Artificial Intelligence (IJCAI)}. 1776--1782.
\newblock


\bibitem[\protect\citeauthoryear{Young and Ward}{Young and Ward}{2001}]%
        {young01}
{J.~M. Young} {and} {J.~E. Ward}. 2001.
\newblock \showarticletitle{Evidence-based Medicine in General Practice:
  Beliefs and Barriers Among Australian GPs}.
\newblock {\em Journal of Evaluation in Clinical Practice\/} {7}, 2 (2001),
  201--210.
\newblock


\bibitem[\protect\citeauthoryear{Yu and Cao}{Yu and Cao}{2008}]%
        {yu08}
{H. Yu} {and} {Y. Cao}. 2008.
\newblock \showarticletitle{Automatically Extracting Information Needs from Ad
  Hoc Clinical Questions}. In {\em {Proceedings of AMIA Annual Symposium}}.
  96--100.
\newblock


\bibitem[\protect\citeauthoryear{Yu and Kaufman}{Yu and Kaufman}{2007}]%
        {yu07b}
{H. Yu} {and} {D. Kaufman}. 2007.
\newblock \showarticletitle{A Cognitive Evaluation of Four Online Search
  Engines for Answering Definitional Questions Posed by Physicians}.
\newblock {\em Pacific Symposium on Biocomputing\/}  {12} (2007), 328--339.
\newblock


\bibitem[\protect\citeauthoryear{Yu, Lee, Kaufman, Ely, Osheroff, Hripcsak, and
  Cimino}{Yu et~al\mbox{.}}{2007}]%
        {yu07a}
{H. Yu}, {M. Lee}, {D. Kaufman}, {J. Ely}, {J.~A. Osheroff}, {G. Hripcsak},
  {and} {J. Cimino}. 2007.
\newblock \showarticletitle{Development, implementation and, a cognitive
  evaluation of a definitional question answering system for physicians}.
\newblock {\em Journal of Biomedical Informatics\/}  {40} (2007), 236--251.
\newblock


\bibitem[\protect\citeauthoryear{Yu and Sable}{Yu and Sable}{2005}]%
        {yu05a}
{H. Yu} {and} {C. Sable}. 2005.
\newblock \showarticletitle{{Being Erlan Shen: Identifying Answerable
  Questions}}. In {\em Proceedings of the IJCAI'05 Workshop on Knowledge and
  Reasoning for Answering Questions (KRAQ'05)}. 6--14.
\newblock


\bibitem[\protect\citeauthoryear{Yu, Sable, and Zhu}{Yu et~al\mbox{.}}{2005}]%
        {yu05b}
{H. Yu}, {C. Sable}, {and} {H. Zhu}. 2005.
\newblock \showarticletitle{Classifying Medical Questions based on an Evidence
  Taxonomy}. In {\em {Proceedings of the AAAI Workshop Question Answering in
  Restricted Domains}}. 27--35.
\newblock


\bibitem[\protect\citeauthoryear{Zeng, Goryachev, Weiss, Sordo, Myurphy, and
  Lazarus}{Zeng et~al\mbox{.}}{2006}]%
        {zeng06}
{Q.~T. Zeng}, {S. Goryachev}, {S. Weiss}, {M. Sordo}, {S.~N. Myurphy}, {and}
  {R. Lazarus}. 2006.
\newblock \showarticletitle{Extracting principal diagnosis, co-morbidity and
  smoking status for asthma research: evaluation of a natural language
  processing system}.
\newblock {\em {BMC Medical Informatics and Decision Making}\/}  {6} (2006),
  30--38.
\newblock


\bibitem[\protect\citeauthoryear{Zhou, Lin, Munteanu, and Hovy}{Zhou
  et~al\mbox{.}}{2006}]%
        {zhou06}
{L. Zhou}, {C. Lin}, {D.~S. Munteanu}, {and} {E. Hovy}. 2006.
\newblock \showarticletitle{ParaEval: Using Paraphrases to Evaluate Summaries
  Automatically}. In {\em Proceedings of NAACL-HLT}. 447--454.
\newblock


\bibitem[\protect\citeauthoryear{Zweigenbaum}{Zweigenbaum}{2003}]%
        {zweigenbaum03}
{P. Zweigenbaum}. 2003.
\newblock \showarticletitle{Question Answering in Biomedicine}. In {\em
  Proceedings of the EACL 2003 Workshop on Natural Language Processing for
  Question Answering}.
\newblock


\bibitem[\protect\citeauthoryear{Zweigenbaum}{Zweigenbaum}{2009}]%
        {zweigenbaum09}
{P. Zweigenbaum}. 2009.
\newblock \showarticletitle{{Knowledge and reasoning for medical
  question-answering}}. In {\em Proceedings of the 2009 Workshop on Knowledge
  and Reasoning for Answering Questions, ACL-IJCNLP 2009}. 1--2.
\newblock


\bibitem[\protect\citeauthoryear{Zweigenbaum, Demner-Fushman, Yu, and
  Cohen}{Zweigenbaum et~al\mbox{.}}{2007}]%
        {zweigenbaum07}
{P. Zweigenbaum}, {D. Demner-Fushman}, {H. Yu}, {and} {K.~B. Cohen}. 2007.
\newblock \showarticletitle{Frontiers of biomedical text mining: current
  progress}.
\newblock {\em Briefings in Bioinformatics\/} {8}, 5 (September 2007),
  358--375.
\newblock


\end{thebibliography}

\received{..}{..}{..}


\end{document}